%% file: IFNO.tex
\documentclass[twoside]{article}
\usepackage[accepted]{aistats2025}

\usepackage[utf8]{inputenc} % allow utf-8 input
\usepackage[T1]{fontenc}    % use 8-bit T1 fonts
\usepackage{hyperref}       % hyperlinks
\usepackage{url}            % simple URL typesettings
\usepackage{booktabs}       % professional-quality tables
\usepackage{amsfonts}       % blackboard math symbols
\usepackage{nicefrac}       % compact symbols for 1/2, etc.
\usepackage{microtype}      % microtypography

\usepackage{microtype}
\usepackage{graphicx}
\usepackage{natbib}
\usepackage{amssymb, amsmath,amsthm}
\usepackage{amsfonts}
\usepackage{algorithm}
\usepackage{algorithmic}
\usepackage{paralist}
\usepackage{multirow}
\usepackage{times}
\usepackage{url,enumerate}
\usepackage{color,xcolor}
\usepackage{makeidx}  % allows for indexgeneration
\usepackage{amsmath,amssymb}
\usepackage{mathtools}
\usepackage[small, compact]{titlesec}
\usepackage{xspace}
\usepackage{epstopdf}
\usepackage{mathrsfs}
\usepackage{times}
\usepackage{enumerate}
\usepackage{color}
\usepackage{graphicx,epsfig}
\usepackage{amsmath,amssymb,xspace}
\usepackage{url}
\usepackage{bm}
\usepackage{bbm}
\usepackage{upgreek}
\usepackage{cleveref}
\usepackage{multirow}
\usepackage{ulem}
\usepackage{cancel}
\usepackage{empheq}
\usepackage{pifont}
\usepackage{subcaption}
\usepackage{amsmath}
\usepackage{amssymb}
\usepackage{mathtools}
\usepackage{amsthm}

\theoremstyle{plain}

\theoremstyle{definition}

\theoremstyle{remark}

\usepackage{xcolor}

\begin{document}

\newcommand{\ours}{iFNO\xspace}
\newcommand{\IDON}{iDON\xspace}
\newcommand{\nio}{NIO\xspace}
\newcommand{\clas}{IOFM\xspace}


\twocolumn[

\aistatstitle{Invertible Fourier Neural Operators for Tackling Both Forward and Inverse Problems}

\aistatsauthor{Da Long \And Zhitong Xu \And  Qiwei Yuan  \And Yin Yang \And Shandian Zhe}

%\aistatsaddress{ University of Utah \And University of Utah \And University of Utah \And University of Utah \And University of Utah } ]
\aistatsaddress{ Kahlert School of Computing, University of Utah\\
\texttt{\{da.long, u1502956, joshua.yuan, yin.yang\}@utah.edu,  zhe@cs.utah.edu}} ]

\input{./emacscomm.tex}
\input{./abstract}
\input{./intro}

\input{./model}
\input{./related}
\input{./exp}

\input{./conclusion}

\bibliographystyle{apalike}
\bibliography{IFNO}

\newpage
\appendix
\onecolumn
\input{appendix.tex}
\clearpage

\end{document}

%% file: emacscomm.tex
% Math commands by Thomas Minka
\newcommand{\var}{{\rm var}}
\newcommand{\Tr}{^{\rm T}}
\newcommand{\vtrans}[2]{{#1}^{(#2)}}
\newcommand{\kron}{\otimes}
\newcommand{\schur}[2]{({#1} | {#2})}
\newcommand{\schurdet}[2]{\left| ({#1} | {#2}) \right|}
\newcommand{\had}{\circ}
\newcommand{\diag}{{\rm diag}}
\newcommand{\invdiag}{\diag^{-1}}
\newcommand{\rank}{{\rm rank}}
% careful: ``null'' is already a latex command
\newcommand{\nullsp}{{\rm null}}
\newcommand{\tr}{{\rm tr}}
\renewcommand{\vec}{{\rm vec}}
\newcommand{\vech}{{\rm vech}}
\renewcommand{\det}[1]{\left| #1 \right|}
\newcommand{\pdet}[1]{\left| #1 \right|_{+}}
\newcommand{\pinv}[1]{#1^{+}}
\newcommand{\erf}{{\rm erf}}
\newcommand{\hypergeom}[2]{{}_{#1}F_{#2}}
\newcommand{\tka}{{\tilde{\kappa} }}
% boldface characters
\renewcommand{\a}{{\bf a}}
\renewcommand{\b}{{\bf b}}
\renewcommand{\c}{{\bf c}}
\renewcommand{\d}{{\rm d}}  % for derivatives
\newcommand{\e}{{\bf e}}
\newcommand{\f}{{\bf f}}
\newcommand{\g}{{\bf g}}
\newcommand{\h}{{\bf h}}
%\newcommand{\k}{{\bf k}}
% in Latex2e this must be renewcommand
\renewcommand{\k}{{\bf k}}
\newcommand{\m}{{\bf m}}
\newcommand{\mb}{{\bf m}}
\newcommand{\n}{{\bf n}}
\renewcommand{\o}{{\bf o}}
\newcommand{\p}{{\bf p}}
\newcommand{\q}{{\bf q}}
\renewcommand{\r}{{\bf r}}
\newcommand{\s}{{\bf s}}
\renewcommand{\t}{{\bf t}}
\renewcommand{\u}{{\bf u}}
\renewcommand{\v}{{\bf v}}
\newcommand{\w}{{\bf w}}
\newcommand{\x}{{\bf x}}
\newcommand{\hx}{{\hat{\x}}}
\newcommand{\hf}{{\hat{\f}}}

\newcommand{\y}{{\bf y}}
\newcommand{\z}{{\bf z}}
%s\newcommand{\l}{\boldsymbol{l}}
\newcommand{\bepsi}{{\boldsymbol \epsilon}}
\newcommand{\A}{{\bf A}}
\newcommand{\B}{{\bf B}}
\newcommand{\C}{{\bf C}}
\newcommand{\D}{{\bf D}}
\newcommand{\E}{{\bf E}}
\newcommand{\F}{{\bf F}}
\newcommand{\G}{{\bf G}}
\renewcommand{\H}{{\bf H}}
\newcommand{\I}{{\bf I}}
\newcommand{\J}{{\bf J}}
\newcommand{\K}{{\bf K}}
\newcommand{\hK}{\widehat{\K}}
\renewcommand{\L}{{\bf L}}
\newcommand{\M}{{\bf M}}
\newcommand{\N}{\mathcal{N}}  % for normal density
\newcommand{\gp}{\mathcal{GP}}  % for normal density
\newcommand{\Acal}{\mathcal{A}}
\newcommand{\Ocal}{\mathcal{O}}
\newcommand{\Dcal}{\mathcal{D}}
\newcommand{\Ycal}{\mathcal{Y}}
\newcommand{\Zcal}{\mathcal{Z}}
\newcommand{\Fcal}{\mathcal{F}}
\newcommand{\Vcal}{\mathcal{V}}
\newcommand{\Lcal}{\mathcal{L}}
\newcommand{\Tcal}{\mathcal{T}}
\newcommand{\Gcal}{\mathcal{G}}
\newcommand{\Hcal}{\mathcal{H}}
\newcommand{\Scal}{\mathcal{S}}
\newcommand{\Pcal}{\mathcal{P}}
\newcommand{\ifb}{\mathcal{IF}}

\renewcommand{\O}{{\bf O}}
\renewcommand{\P}{{\bf P}}
\newcommand{\Q}{{\bf Q}}
\newcommand{\R}{{\bf R}}
\renewcommand{\S}{{\bf S}}
\newcommand{\T}{{\bf T}}
\newcommand{\U}{{\bf U}}
\newcommand{\V}{{\bf V}}
\newcommand{\W}{{\bf W}}
\newcommand{\X}{{\bf X}}
\newcommand{\hX}{{\hat{\X}}}
\newcommand{\Y}{{\bf Y}}
\newcommand{\Z}{{\bf Z}}
\newcommand{\Mcal}{{\mathcal{M}}}
\newcommand{\Wcal}{{\mathcal{W}}}
\newcommand{\Ucal}{{\mathcal{U}}}
\newcommand{\Qcal}{{\mathcal{Q}}}
\newcommand{\Rcal}{{\mathcal{R}}}

\newcommand{\zhat}{{\widehat{\z}}}
\newcommand{\xhat}{{\widehat{x}}}
\newcommand{\that}{{\widehat{t}}}

% this is for latex 2.09
% unfortunately, the result is slanted - use Latex2e instead
%\newcommand{\bfLambda}{\mbox{\boldmath$\Lambda$}}
% this is for Latex2e
\newcommand{\bfLambda}{\boldsymbol{\Lambda}}

% Yuan Qi's boldsymbol
\newcommand{\bsigma}{\boldsymbol{\sigma}}
\newcommand{\balpha}{\boldsymbol{\alpha}}
\newcommand{\bpsi}{\boldsymbol{\psi}}
\newcommand{\bphi}{\boldsymbol{\phi}}
\newcommand{\boldeta}{\boldsymbol{\eta}}
\newcommand{\Beta}{\boldsymbol{\eta}}
\newcommand{\btau}{\boldsymbol{\tau}}
\newcommand{\bvarphi}{\boldsymbol{\varphi}}
\newcommand{\bzeta}{\boldsymbol{\zeta}}

\newcommand{\blambda}{\boldsymbol{\lambda}}
\newcommand{\bLambda}{\mathbf{\Lambda}}
\newcommand{\bOmega}{\mathbf{\Omega}}
\newcommand{\bomega}{\mathbf{\omega}}
\newcommand{\bPi}{\mathbf{\Pi}}

\newcommand{\cov}{\text{cov}}
\newcommand{\bpi}{\boldsymbol{\pi}}
\newcommand{\btheta}{\boldsymbol{\theta}}
\newcommand{\bxi}{\boldsymbol{\xi}}
\newcommand{\bSigma}{\boldsymbol{\Sigma}}

\newcommand{\bgamma}{\boldsymbol{\gamma}}
\newcommand{\bGamma}{\mathbf{\Gamma}}

\newcommand{\bmu}{\boldsymbol{\mu}}
\newcommand{\1}{{\bf 1}}
\newcommand{\0}{{\bf 0}}

\newcommand{\bs}{\backslash}
\newcommand{\ben}{\begin{enumerate}}
\newcommand{\een}{\end{enumerate}}

 \newcommand{\notS}{{\backslash S}}
 \newcommand{\nots}{{\backslash s}}
 \newcommand{\noti}{{\backslash i}}
 \newcommand{\notj}{{\backslash j}}
 \newcommand{\nott}{\backslash t}
 \newcommand{\notone}{{\backslash 1}}
 \newcommand{\nottp}{\backslash t+1}

\newcommand{\notk}{{^{\backslash k}}}
\newcommand{\notij}{{^{\backslash i,j}}}
\newcommand{\notg}{{^{\backslash g}}}
\newcommand{\wnoti}{{_{\w}^{\backslash i}}}
\newcommand{\wnotg}{{_{\w}^{\backslash g}}}
\newcommand{\vnotij}{{_{\v}^{\backslash i,j}}}
\newcommand{\vnotg}{{_{\v}^{\backslash g}}}
\newcommand{\half}{\frac{1}{2}}
\newcommand{\msgb}{m_{t \leftarrow t+1}}
\newcommand{\msgf}{m_{t \rightarrow t+1}}
\newcommand{\msgfp}{m_{t-1 \rightarrow t}}

\newcommand{\proj}[1]{{\rm proj}\negmedspace\left[#1\right]}
\newcommand{\argmin}{\operatornamewithlimits{argmin}}
\newcommand{\argmax}{\operatornamewithlimits{argmax}}

\newcommand{\dif}{\mathrm{d}}
\newcommand{\abs}[1]{\lvert#1\rvert}
\newcommand{\norm}[1]{\lVert#1\rVert}
\newcommand{\hu}{\widehat{\u}}
%miscellaneous symbols
%\newcommand{\ie}{{{\em i.e.,}}\xspace}
\newcommand{\ie}{{\textit{i.e.,}}\xspace}
\newcommand{\eg}{{\textit{e.g.,}}\xspace}
\newcommand{\etc}{{\textit{etc.}}\xspace}
\newcommand{\EE}{\mathbb{E}}
\newcommand{\dr}[1]{\nabla #1}
\newcommand{\VV}{\mathbb{V}}
\newcommand{\sbr}[1]{\left[#1\right]}
\newcommand{\rbr}[1]{\left(#1\right)}
\newcommand{\cmt}[1]{}
\newcommand{\tg}{\widetilde{g}}
\newcommand{\tZ}{\widetilde{\Z}}
\newcommand{\teps}{\widetilde{\epsilon}}

\newcommand{\bi}{{\bf i}}
\newcommand{\bj}{{\bf j}}
\newcommand{\bK}{{\bf K}}

\newcommand{\unit}[1]{\,\mathrm{#1}}

%% file: abstract.tex
%Operator learning --> muti-reoslution --> ....
\begin{abstract}
Fourier Neural Operator (FNO) is  a powerful and popular operator learning method. However, FNO is mainly used in forward prediction, yet a great many applications rely on solving inverse problems. In this paper, we propose an invertible Fourier Neural Operator (\ours) for jointly tackling the forward and inverse problems. We developed a series of invertible Fourier blocks in the latent channel space to share the model parameters, exchange the information, and mutually regularize the learning for the bi-directional tasks. We integrated a variational auto-encoder to capture the intrinsic structures within the input space and to enable posterior inference so as to mitigate challenges of illposedness, data shortage, noises that are common in inverse problems. We proposed a three-step process to combine the invertible blocks and the VAE component for effective training. The evaluations on seven benchmark forward and inverse tasks have demonstrated the advantages of our approach. The code is available at~\url{https://github.com/BayesianAIGroup/iFNO}.
\end{abstract}

%% file: intro.tex
\section{INTRODUCTION}
Operator learning (OL) is currently at the forefront of AI for science. It seeks to estimate function-to-function mappings from data and can be used as a valuable surrogate in various applications related to scientific simulation. Among the notable approaches in this domain is the Fourier neural operator (FNO)~\citep{li2020fourier}. Leveraging the convolution theorem and fast Fourier transform (FFT), FNO executes a sequence of global linear transform and nonlinear activation within the functional space to capture complex function-to-function mappings. FNO is computationally  efficient, and has shown excellent performance across many OL tasks.

Despite many success stories~\citep{pathak2022fourcastnet,kovachki2023neural,kashefi2024novel}, FNO is primarily applied for solving forward problems. Typically, it is used to predict solution functions based on the input sources, parameters of partial differential equations (PDEs), and/or initial conditions. However, practical applications often involve another crucial category of tasks, namely inverse problems~\citep{stuart2010inverse}. For instance, given the solution measurements, how to deduce the unknown sources, how to determine the PDE parameters, or identify the initial conditions? Inverse problems tend to be more challenging due to issues such as ill-posedness~\citep{tikhonov1963solution,engl2014inverse}, noisy measurements, and limited data quantity. Even if one attempts to directly train an FNO to map the measured solution function back to the input of interest, the aforementioned challenges persist.

%inverstable block in the lifted channel space --- lifting and project for u-->f
To bridge this gap, we propose \ours, a novel invertible Fourier neural operator jointly addressing both forward and inverse problems. %while providing posterior estimation and uncertainty quantification. 
Due to the sharing of the model parameters, 
the co-learning for the bi-directional tasks enables efficient information exchange and mutual regularization, enhancing performance on both fronts and mitigating challenges especially for inverse problems. 
Specifically, we first develop a series of invertible Fourier blocks in the lifted channel space, capturing rich representation information. Each block takes a pair of inputs with an equal number of channels, generating outputs through the Fourier layer of FNO, softplus transform, and element-wise multiplication. This ensures a rigorous bijection pair between the inputs and outputs during expressive functional transforms in the latent channel space. To enable inverse prediction in the original space, we incorporate a pair of multi-layer perceptions (MLPs) that lift the final prediction's channel back to the latent space and project the latent channels to the input space. Second, to capture intrinsic structures within the input space and further mitigate challenges like ill-posedness and data shortage, we introduce a low-dimensional representation and integrate a variational auto-encoder (VAE) to reconstruct the input function values. The VAE component also enables generation of posterior samples for prediction. Third, for effective training, we developed a three-step process: We first train the invertible blocks and the VAE component separately, then combine them to continue training, resulting in the final model.

For evaluation, we examined our method in seven benchmark problems, including scenarios based on Darcy flow, second-order wave propagation, diffusion reaction, and Naiver-Stoke (NS) equations.  In addition to forward solution prediction, our method was tested for inverse inference, specifically deducing permeability, square slowness, initial conditions, earlier system states from (noisy) solution or system state measurements. 
We compared with the recent inverse neural operator~\citep{molinaro2023neural} designed explicitly for solving inverse problems, and invertible deep operator net~\citep{kaltenbach2022semi}. Comparisons were also made with FNO and another recent state-of-the-part attention-based neural operator, both of which were trained \textit{separately} for forward and inverse prediction. In addition, we tested with adapting FNO to the classical framework for solving inverse problem. That is, one optimizes the input to the forward model to match the prediction with observations. 
Across all the tasks, \ours consistently delivers the best or near-best performance, often outperforming the competing methods by a large margin. In the vast majority of cases, \ours significantly improves upon the standard FNO in both forward and inverse predictions. 
Visualization of the predictive variance for the inverse problems reveals intriguing and reasonable uncertainty calibration results.

%% file: model.tex
\section{PRELIMINARIES}

\textbf{Operator Learning}. Consider learning a mapping between two function spaces (\eg Banach spaces) $\psi: \Hcal\rightarrow\Ucal$. We collect a training dataset that consists of pairs of discretized input and output functions, denoted by $\mathcal{D}=\{(\f_n,\u_n)\}_{n=1}^N$. Each $\f_n$ and $\u_n$ represents samples from an input function
$f_n\in \Hcal$ and the corresponding output function $u_n = \psi(f_n) \in \Ucal$, respectively. Both $f_n$ and $u_n$ are sampled at evenly-spaced locations, \eg over a 64 $\times$ 64 grid in $[0, 1]^2$.%the 2D domain $[0,1]\times [0, 1]$.

\textbf{Fourier Neural Operator (FNO)} first lifts the sampled input function values $\f$ into a higher-dimensional feature space (via an MLP) to enrich the representation. Then it uses a series of Fourier layers to alternatingly conduct linear transforms and nonlinear activation, %$v_{t+1}(\x) \leftarrow \sigma\left(\Wcal v_t(\x) + \int \kappa(\x - \x')v_t(\x') \d \x' \right)$, 
 \begin{align}
 	v_{t+1}(\x) \leftarrow \sigma\left(\Wcal v_t(\x) + \int \kappa(\x - \x')v_t(\x') \d \x' \right), \label{eq:fourier-layer} 
 \end{align}
where $v_t$ is the input to the $t$-th layer, $v_{t+1}$ the output, $\kappa$ the integration kernel, and $\sigma$ the activation. Utilizing the convolution theorem expressed as $$ \int \kappa(\x - \x')v_t(\x') \d \x' = \Fcal^{-1}\left[\Fcal[\kappa]\cdot\Fcal[v_t]\right](\x),$$ where $\Fcal$ and $\Fcal^{-1}$ denote the Fourier and inverse Fourier transforms, respectively, the Fourier layer first executes a fast Fourier transform (FFT) over $v_t$,  then multiplies the result with the discretized representation of $\kappa$ in the frequency domain, \ie $\Fcal[\kappa]$, and executes inverse FFT; $\mathcal{W}v_t(\mathbf{x})$ perform a location-wise linear transform.  Owning to FFT, the Fourier layer is computationally  efficient. The storage of $\Fcal[\kappa]$ can be memory intensive. In practice, one often truncates the high frequency modes to save the memory cost. After several Fourier layers, another MLP is used to project back from the latent space to generate the final prediction.
The training is typically carried out by minimizing a relative $L_2$ loss.
% , 
% \begin{align}
%     \Theta^* = \argmin_{\Theta}\frac{1}{N}\sum_{n=1}^N \frac{ \|\u_n - \psi_{\text{FNO}}(\f_n;\Theta)\|}{\|\u_n\|},
% \end{align}
% where $\|\cdot\|$ is often chosen as the Frobenius norm, and $\Theta$ includes all the model parameters. 
\begin{figure*}[t]
	\centering
	\setlength\tabcolsep{0pt}
 \includegraphics[width=\textwidth]{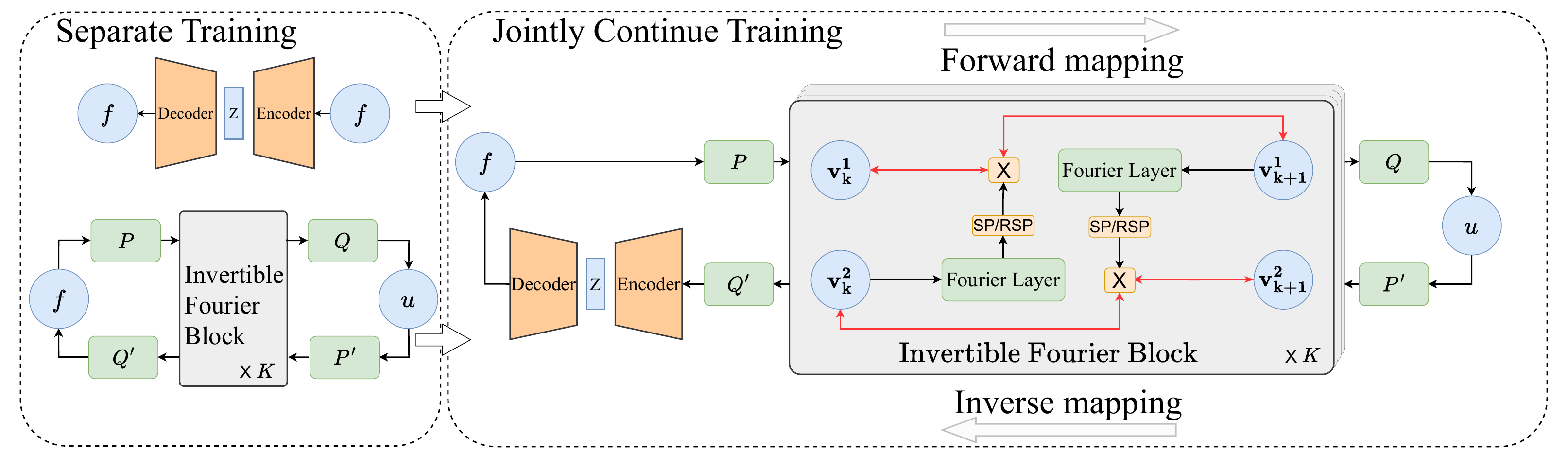}
	\caption{\small An Overview of \ours. Left panel: the $\beta$-VAE module and the invertible Fourier blocks. Right panel: the entire architecture. ``SP'' and ``RSP'' denote softplus and the reciprocal of the softplus, respectively. We first train separately the $\beta$-VAE module and invertible Fourier blocks as shown in the first panel. Then we combine them to continue training the entire model as depicted in the right panel.} \label{fig:architecture}
\end{figure*}

\section{INVERTIBLE FOURIER NEURAL OPERATORS}
Neural operators, including FNO, are commonly employed to address forward problems. %, namely predicting or computing the outcomes based on given inputs to the system of interest. 
Taking PDE systems as an example, the input  typically comprises external sources or forces, system parameters, and/or initial conditions, with the output representing the corresponding solution function. However, many applications require deducing unknown causes from the observed effect or measurement data, a scenario known as inverse problems. For instance, this involves inferring unknown sources, system parameters, or initial conditions from solution measurements. 
Inverse problems have broad significance in scientific and engineering domains, \eg ~\citep{tanaka1998inverse,yilmaz2001seismic, nashed2002inverse}. However, tackling inverse problems is notably challenging due to their inherently ill-posed nature~\citep{tikhonov1963solution,stuart2010inverse,engl2014inverse}. In contrast to forward problems, inverse problems often lack a unique solution and are highly sensitive to variations in data. These challenges are further compounded in complex applications where measurement data is often limited, noisy, and/or inaccurate.

To better address these challenges, we propose iFNO, an invertible Fourier neural operator jointly tackling both the forward and inverse problems. Gven the training dataset $\Dcal = \{\f_n, \u_n\}_{n=1}^N$, our goal is to jointly learn the forward and inverse mappings via a \textit{unified} neural operator. 
% \begin{align}
%     \psi:\Hcal \rightarrow \Ucal, \;\;\;\; \psi^{-1}: \Ucal \rightarrow \Hcal.
% \end{align}
By sharing the model parameters across both tasks, we conduct 
co-learning for the bi-directional tasks to enable efficient information exchange and mutual regularization, so as to improve performance on both fronts and alleviate aforementioned challenges like ill-posedness and data noise. The details of \ours are specified as follows. 

\subsection{Invertible Fourier Blocks in Latent Space} 
To enable invertible computing and prediction while preserving the expressiveness of the original FNO, we first design and stack a series of invertible Fourier blocks in the latent channel space. Specifically, following the standard FNO, we  apply a multi-layer preceptron (MLP) $\Pcal$ over each element of the discretized input function $\f$ and the sampling locations to map $\f$ into a higher-dimensional latent space with $2d$ channels. That means, at each sampling location, we have a $2d$-dimensional feature representation.  We split the channels of $\Pcal(\f)$ into halves, each with $d$ channels, and feed them into a series of invertible Fourier blocks $\ifb_1, \ldots, \ifb_K$. Each block $\ifb_k$ receives a pair of inputs $\v^1_k$ and $\v^2_k$, and produces a pair of outputs $\v^1_{k+1}$ and $\v^2_{k+1}$, which are then fed into  $\ifb_{k+1}$. All the input and outputs possess the same number  of channels. We use the framework in \citep{dinh2016density} to design each block $\ifb_k$,
\begin{align}
\v_{k+1}^1 &\leftarrow \v_k^1\odot S(\mathcal{L}(\v_k^2)), \notag \\
\v_{k+1}^2 &\leftarrow \v_k^2\odot S(\mathcal{L}(\v_{k+1}^1)), \label{eq:ifb}
\end{align}
where $\odot$ is the element-wise multiplication, $\Lcal$  is the Fourier layer of the standard FNO that fulfills the linear transform and nonlinear activation as expressed in \eqref{eq:fourier-layer}, and $S$ is the element-wise softplus transform\footnote{We did not use the exponential transform as suggested in~\citep{dinh2016density}. We empirically found that the softplus transform is numerically more stable and consistently achieved better performance. }, $S(x) = \tau^{-1}\log \left(1+\exp(\tau x)\right)$, 
% \begin{align}
%     S(x) = \frac{1}{\tau}\log \left(1+\exp(\tau x)\right),
%     \label{eq:sp}
% \end{align}
where $\tau$ is a hyperparameter to adjust the shape. 
In this way, $\{\v^1_k, \v^2_k\}$ and $\{\v^1_{k+1}, \v^2_{k+1}\}$ form a bijection pair. Given the outputs $\{\v_{k+1}^1,\v_{k+1}^2\}$, the inputs can be inversely computed via
\begin{align} 
\v_k^2 &\leftarrow \v_{k+1}^2\odot \left[S(\Lcal(\v_{k+1}^1))\right]^{-1},\notag \\
\v_k^1 &\leftarrow \v_{k+1}^1\odot \left[S(\Lcal(\v_{k}^2))\right]^{-1}, \label{eq:invertible-block-backward}
\end{align}
where $[\cdot]^{-1}$ denotes the element-wise reciprocal. 

The outputs of the last invertible Fourier block, $\v^1_K$ and $\v^2_K$, are concatenated and fed into a second MLP $\Qcal$, which projects back to generate the prediction of the output function at the sampling locations, 
\begin{align}
\psi_{\text{\ours}}(\f) = \Qcal(\v^1_K, \v^2_K). \label{eq:fwd-pred}
\end{align}

Next, to fulfill the inverse prediction from the (discretized) output function $\u$ back to the input function, we introduce another pairs of MLPs, $\Pcal'$ and $\Qcal'$, for which, $\Pcal'$ first maps $\u$ back to the latent space to predict the outputs of the last invertible Fourier block,
\[
\v^1_K, \v^2_K = \Pcal'(\u),
\]
then inversion \eqref{eq:invertible-block-backward} is sequentially executed to predict the inputs to each block until the first one, and finally $\Q'$ projects the inputs to the first invertible Fourier block back to generate the prediction of the original input function, 
\begin{align}
\psi^{-1}_{\text{\ours}}(\u) = \Q'(\v^1_1, \v^2_1). \label{eq:inv-pred}
\end{align}
One might question why not position the invertible Fourier blocks directly in the original space, eliminating channel lifting and projection. Adopting this approach will  limit the representation power and miss the rich information in the higher-dimensional latent space. Additionally, deciding whether to split the original input function samples into halves or duplicate them into two copies becomes a challenge. The former can potentially disrupt the internal structures of the input function, while the latter introduces additional training issues, such as how to enforce the prediction of the two inputs to be identical. We found that empirically in both our model and  standard FNO, channel lifting is crucial to achieve promising performance.
%One might question why not situate the invertible Fourier blocks in the original space, eliminating the channel lifting and projection. However, adopting this approach will severely restrict the representation power, and miss the rich information in the higher dimensional latent space. It also becomes challenging to decide whether to split the original input function into halves or simply duplicate it into two copies. The former may potentially disrupt the internal structures of the input function, while the latter introduces additional training issues, \eg how to enforce the prediction of the two inputs to be identical. We found that, in both our model and the standard FNO, channel lifting and projection is an important step to achieve promising performance. 

\cmt{
To create a fully invertible neural operator, we utilize the reversible block RealNVP proposed by \citet{dinh2016density}. With that we can create invertible Fourier blocks to perform successive linear and nonlinear invertible transformations in the functional space. However, the reversible blocks require the dimension of input to be the same as the dimension of output. While for a standard Fourier layer, the input is a higher dimensional representation lifted by a MLP. It would limit the representation capacity if the the reversible block operates solely on Fourier layers receiving low dimensional inputs. On the other hand, the reversible block dividing the input into two halves would disrupt the intrinsic structure of the PDE parameters, boundary conditions, and solutions, etc.

Hence, we propose to build the invertible Fourier blocks in a high dimensional space and the division operates on the lifted dimension to preserve the intrinsic structure. In order to preserve the model's capacity while ensuring an invertible structure, \ours incorporates four MLPs at the start $(P_1,Q_1)$ and the end $(P_2,Q_2)$ of the model. Specifically, $P_1$ takes the input function $\f$ and projects it into a high dimensional representation $\v_0$. Then it progresses through $N$ invertible Fourier blocks $\{F_k\}_{k=1}^{N}$, with each block respectively outputs $\v_1,\v_2,...,$ and $\v_N$. Then $P_2$ projects $\v_N$ back to the target dimension and predicts $\u$. On the backward path, we propose to use a MLP $Q_2$ to recover $\u$ from the output space to a high dimensional space. Similarly, a MLP $Q_1$ is used to recover a high dimensional space representation back to the inverse solution $\f$. To make sure $P_1$ and $Q_1$, $P_2$ and $Q_2$ are two invertible mappings, we propose a reconstruction loss as follows,
\begin{equation} \label{eq:reconstruction-loss}
\small
loss_{\small rec} = \frac{1}{N}\sum_{k=1}^{N} (\norm{Q_1(P_1(\f_k))-\f_k}_2 + \norm{P_2(Q_2(\u_k))-\u_k}_2)
\end{equation}

In this way, invertibility is preserved while $\f$ and $\u$ can be arbitrary dimensional as long as we lift them into the same dimension, and we can control the model capacity by adjusting the "lifting dimension". Hence, for the backward path, the PDE solution $\u$ is lifted to a high dimensional representation $\z_N$ by $Q_2$, which has the same dimension as $\v_N$. Next, $\z_N$ passes through $\{F_k\}_{k=1}^{4}$ reversibly and iteratively. At last, $Q_1$ maps $z_1$ back to the input space and get predicted $\f$. 
Specifically, each invertible Fourier block's input $\v_k$ is split into two halves, $\v_k^1$ and $\v_k^2$ over the dimension that results from dimensional lift. For instance, if $\v_k$ is lifted into a 64-dimensional hidden representation, then $\v_k^1$ and $\v_k^2$ are both 32-dimensional hidden representations. Next, $\v_k^1$ and $\v_k^2$ are transformed by
\begin{equation} \label{eq:invertible-block-forward}
\begin{split}
\v_{k+1}^1=&\v_k^1\odot t(\mathcal{L}_2(\v_k^2)) \\
\v_{k+1}^2=&\v_k^2\odot t(\mathcal{L}_1(\v_{k+1}^1))
\end{split}
\end{equation}
where $\{\v_{k+1}^i\}_{i=1}^2$ are the outputs from the kth invertible Fourier block (also the inputs to the (k+1)th invertible Fourier block), $\odot$, $+$ the element-wise multiplication and addition, and $\mathcal{L}$ a standard Fourier layer. Given the output $\v_{k+1}^1$ and $\v_{k+1}^2$, $\v_k^1$ and $\v_k^2$ can be exactly recovered by
\begin{equation} \label{eq:invertible-block-backward}
\begin{split}
\v_k^2=&\v_{k+1}^2\odot \frac{1}{t(\mathcal{L}_1(\v_{k+1}^1))}\\
\quad \v_k^1=&\v_{k+1}^1\odot \frac{1}{t(\mathcal{L}_2(\v_{k}^2))}
\end{split}
\end{equation}
In implementation, we use the Softplus function $t(x)=\frac{1}{\beta}\log (1+\exp(\beta x))$ as a substitute for the exponential function in the original paper. Note that we did not use 
}

% we discard $k_1(u^2)$ and $k_2(v^1)$, and to ensure numerical stability,

\subsection{Embedding Intrinsic Structures}
To extract the intrinsic structures within the input space so as to further mitigate challenges for solving inverse problems, we integrate a $\beta$-variational auto-encoder ($\beta$-VAE)~\citep{higgins2016beta} into our model. Specifically, after projecting $\{\v^1_1, \v^2_1\}$ back to the input space --- see \eqref{eq:inv-pred} --- we feed the results to the an encoder network to obtain a stochastic latent embedding $\z$. Then through a decoder network, we produce the prediction of the input function values. We henceforth modify \eqref{eq:inv-pred} as the follows, 
\begin{align}
    \z &= \text{Encoder}\left(\Qcal'(\v^1_1, \v^1_2), \bepsi\right), \;\; \bepsi \sim \N(\0, \I), \notag \\
    \f &=\psi^{-1}_{\text{\ours}}(\u) =  \text{Decoder}\left(\z\right). \label{eq:vae}
\end{align}
We leverage the auto-encoding variational Bayes framework~\citep{kingma2013auto} to estimate the posterior distribution of $\z$. This allows us to generate the posterior samples of the  prediction for $\f$, enabling the evaluation of the uncertainty. Our overall model is illustrated in Fig. \ref{fig:architecture}.
\cmt{
To further alleviate the ill-posedness with inverse problems, we propose to use a $\beta$-VAE block \citep{higgins2016beta} to capture the target function $f$ in a low dimensional space as well as to measure uncertainties. The encoder applies five convolutional layers to increasingly lift the number of channels to 32, 64, 128, 256, and 512. The decoder first applies five transposed convolutional layers to reduce the number of channels to 512, 256, 128, 64, and 32. Then an output layer consisting of a transposed convolution operator and a convolution operator is applied to output the predicted $f$. The variational lower bound is given by
\begin{equation} \label{eq:VAE-pretrain}
\small
\begin{split}
ELBO =\mathbb{E}_{\z\sim q_{\phi(\z|\f)}}\log p_{\theta}(\f|\z) - \beta \textit{D}_{\textit{KL}}(q_\phi(\z|\f)||p(\z))
\end{split}
\end{equation}

% _{\small \beta-VAE}
where the multiplier $\beta$ serves as a regularization hyper-parameter, constraining the latent information capacity to balance the complexity of the model and its effectiveness in capturing essential features.
}
\subsection{Three-Step Training}
For effective learning, we perform three steps. We first train the invertible Fourier blocks to fully capture the supervised information (\ie  using~\eqref{eq:inv-pred}; without VAE). The loss is: 
%without the VAE component (\ie removing \eqref{eq:vae} and using \eqref{eq:inv-pred}). The loss is given by 
\begin{align}
    J_\text{IFB} = J_\text{FWD} + J_\text{INV} + J_{\Pcal,\Qcal'} + J_{\Pcal',\Qcal},
\end{align}
where 
\begin{align}
    J_{\text{FWD}} &= \frac{1}{N}\sum\nolimits_{n=1}^N \frac{ \|\u_n - \psi_{\text{\ours}}(\f_n)\|}{\|\u_n\|},  \notag \\
    J_{\text{INV}} &= \frac{1}{N}\sum\nolimits_{n=1}^N  \frac{ \|\f_n - \psi^{-1}_{\text{\ours}}(\u_n)\|}{\|\f_n\|}
\end{align}
measure the data fitness for forward and inverse predictions, $\|\cdot\|$ is the Frobenius norm,  and
\begin{align}
    J_{\Pcal,\Qcal'} &= \frac{1}{N} \sum_{n=1}^N \frac{\|\f_n - \Qcal'(\Pcal(\f_n)) \|}{\|\f_n\|}, \notag \\
    J_{\Pcal', \Qcal} &= \frac{1}{N} \sum_{n=1}^N \frac{\|\u_n - \Qcal(\Pcal'(\u_n))\|}{\|\u_n\|}
\end{align}
are two additional reconstruction loss terms that encourage the invariance of the original information after channel lifting and projection at both the input and output ends.

Next, we train the $\beta$-VAE component to fully capture the hidden structures wthin the input function values.
%exclusively with the input function values. This adaptation of the parameters aims to capture the hidden structures within the input function space. 
We minimize a variational free energy, %$J_{\beta-\text{VAE}} = \frac{1}{N} \sum_{n=1}^N J_n$, where $J_n = \beta\cdot  \text{KL}(q(\z_n) \| p(\z_n)) + \EE_{q}\left[\frac{\|\f_n -\text{Decoder}(\z_n)\|}{\|\f_n\|} \right]$, 
  \begin{align}
      &J_{\beta-\text{VAE}} = \frac{1}{N} \sum_{n=1}^N J_n, \label{eq:beta-vae} \\
      &J_n = \beta  \text{KL}(q(\z_n) \| p(\z_n)) 
      + \EE_{q}\left[\frac{\|\f_n -\text{Decoder}(\z_n)\|}{\|\f_n\|} \right], \notag 
  \end{align}
where $\text{KL}$ is the Kullback-Leibler divergence, $\beta$ is a hyper-parameter, $\z_n$ is the stochastic embedding representation of $\f_n$, and $q(\z_n)$ is the posterior distribution. The samples from $q(\z_n)$ is generated by: $\z_n = \text{Encoder}(\f_n, \bepsi), \;\bepsi\sim \N(\0, \I)$. 
%defined in \eqref{eq:vae}, with the input to the encoder network (that generates the posterior samples of $z_n)$ is $\f_n$.
%We use the reparameterization trick~\citep{kingma2013auto} for stochastic optimization. 

Finally, we combine the two trained models, start with their current parameters, and train the entire model by minimizing a joint loss, 
\begin{align}
    J =  J_\text{FWD} + J_{P,Q'} + J_{P',Q}+ J_{\beta-\text{VAE}},\label{eq:joint-loss}
\end{align}
where according to \eqref{eq:vae}, the inverse prediction is now $\psi^{-1}_{\text{iFNO}}(\u_n) = \text{Decoder}(\z_n)$, and the input to the encoder network for each $\z_n$ is generated by lifting $\u_n$ with $\Pcal'$, going through invertible Fourier blocks, $\ifb_{K}, \ldots, \ifb_1$, and then applying projection MLP $\Qcal'$. See Figure~\ref{fig:architecture}. One might consider directly minimizing the joint loss~\eqref{eq:joint-loss} instead, but this could introduce complications and reduce the efficiency in extracting both supervised and structural knowledge from data. Empirically,  we  found that our tree-step training process results in more stable predictive performance.

\cmt{
The overview of the \ours's architecture is shown in Fig. \ref{fig:architecture}. Overall, $\f$ is lifted to a high dimensional representation by $P_1$, progresses through the forward path of the invertible Fourier blocks, and then gets projected back by $Q_1$ to get the prediction of $\u$. While simultaneously, $\u$ is recovered to a high representation by $Q_2$, progresses through the backward path of the invertible Fourier blocks, and following this, $\u$ it is further processed by the $\beta$-VAE module to get the prediction of $\f$.

%, where the left side is the $\beta$-VAE module and the right side is the \ours module.

For the purpose of efficient training and optimal performance, we first separately pre-train both the \ours module (the bottom module in the Pre-training section) and the $\beta$-VAE module (the top module in the Pre-training section). We pre-train the $\beta$-VAE module in an unsupervised manner as a prior knowledge according to the equation \ref{eq:VAE-pretrain}. The \ours module's loss is defined as follows,
% \begin{equation} \label{eq:pretrain-INFO-loss}

% loss_{\small IFNO} = \sum_{k=1}^{N} (\frac{||(\mathcal{N}_{IFNO}(f_k)-u_k||_2}{||u_k||_2}+\frac{||(\mathcal{N}_{IFNO}^{-1}(u_k)-f_k||_2}{||f_k||_2}) +\\ \frac{1}{N}\sum_{k=1}^{N} (\norm{Q_1(P_1(f_k))-f_k}_2 + \norm{P_2(Q_2(u_k))-u_k}_2) 

% % \mathcal{V}\rightarrow\mathcal{U}
% \end{equation}

\begin{equation} \label{eq:invertible-block-forward}
\small
\begin{split}
loss_{\small IFNO} &=\sum_{k=1}^{N} (\frac{||\mathcal{N}_{IFNO}(\f_k)-\u_k||_2}{||\u_k||_2}+\frac{||\mathcal{N}_{IFNO}^{-1}(\u_k)-\f_k||_2}{||\f_k||_2}) \\ &+  loss_{\small rec}
\end{split}
\end{equation}

Finally, we jointly train the two modules as follows,

\begin{equation} \label{eq:invertible-block-forward}
\small
\begin{split}
loss &= \sum_{k=1}^{N} \frac{||\mathcal{N}_{IFNO}(\f_k)-\u_k||_2}{||\u_k||_2}\\
&-\sum_{k=1}^{N}\mathbb{E}_{\z\sim q_{\phi(\z_k|\f_{k}^{*})}}\log p_{\theta}(\f_k|\z_k)\\
&+\sum_{k=1}^{N}\beta \textit{D}_{\textit{KL}}(q_\phi(\z_k|\f_k)||p(\z_k)+ loss_{\small rec}
\end{split}
\end{equation}

% \mathbb{E}_{\z\sim q_{\phi(z|f)}}\log p_{\theta}(f|z)

% loss_{\small IFNO}
}

%% file: related.tex
\section{RELATED WORK}
Operator learning is a rapidly advancing research field, with various methods falling under the category of neural operators, primarily based on neural networks. Alongside FNO, other notable approaches have been proposed, such as~low-rank neural operator (LNO)~\citep{li2020fourier},  multiwavelet-based NO~\citep{gupta2021multiwavelet}, and convolutional NOs (CNO)~\citep{raonic2024convolutional}. Recently, \citet{li2024multi} proposed active learning methods for multi-resolution FNO. 
Deep Operator Net (DON)~\citep{lu2021learning} is another popular  approach, which consists of a branch net and a trunk net. The branch net is applied over the input function values while the trunk net over the sampling locations. The prediction is generated by the dot product between the outputs of the two nets. To improve the stability and efficiency, ~\citet{lu2022comprehensive} replaced the trunk net by the POD (PCA) bases. Another line of research uses transformers to build NOs~\citep{cao2021choose,li2022transformer,hao2023gnot}. Recent works have also explored kernel operator learning approaches~\citep{long2022kernel, batlle2023kernel}.

\citet{molinaro2023neural} proposed neural inverse operator (NIO), explicitly designed for address inverse problems. NIO sequentially combines DeepONet and FNO, where the DeepONet takes observations and querying locations as the input, and the outputs are subsequently passed to FNO to obtain the prediction for the inverse problems. NIO does not offer uncertainty quantification. 
\citet{kaltenbach2023semi} designed an invertible version of DeepONet. The key idea is to modify the branch net to be invertible following the framework of \citep{dinh2016density}. Given the observations, the approach starts by solving a least-squares problem to restore the output of the branch net, and then proceeds to back-predict the input function values.  The least-squares problem is further casted into a Bayesian inference task and a Gaussian mixture prior is assigned for the unknown output of the branch net. One potential constraint of this approach is that the input and output dimensions of the branch net must be identical for invertibility. When the dimensionality or resolution is high, it can substantially raise the model size and  the computational costs, posing learning challenges (see Sec \ref{sec:experiments}). 

{The classical methods for solving inverse problem is based on the assumption that the forward system is known, \eg a particular PDE system~\citep{stuart2010inverse}. Then one optimizes the input to the forward system to match the system output and measurement data.   Markov-Chain Monte-Carlo (MCMC) sampling is often used to import prior knowledge and to quantify uncertainty. 
}
However, the classical methods require extensively simulating the forward system, such as running a numerical solver, which is very expensive in computation. The recent work~\citep{zhao2022learning} inherits the classical framework, but use a data-driven surrogate model, such as~MeshGraphNet~\citep{pfaff2020learning} and U-Net~\citep{ronneberger2015u}  to replace the forward simulator, so as to accelerate the optimization procedure. Though effective, these approaches need massive simulation examples to train an enough accurate surrogate. Moreover, one has to conduct numerical optimization from scratch for every prediction, which is still quite expensive. See  run time comparison in Appendix Table \ref{tb:run-time}.

%The recent work~\citep{zhao2022learning} aims to tackle the inverse problems with a forward graph neural network simulator (solver), called MeshGraphNet~\citep{pfaff2020learning}. It optimizes the input to the MeshGraphNet to align the output with the solution measurements. While effective, performing numerical optimization for every inverse prediction can be computationally expensive. 
%Moreover, this method relies on knowledge of the underlying system (\eg the order of derivatives in PDEs) for the inverse problem and the ability to simulate the system to generate extensive training examples for MeshGraphNet. However, in practice, such knowledge is often unavailable or inaccurate.

%% file: exp.tex
%small data, visualization large data
%\vspace{-0.1in}

%\vspace{-2mm}
\section{NUMERICAL EXPERIMENTS} 
\label{sec:experiments}
%\vspace{-2mm}
We evaluated our method on seven benchmark problems, each covering both the forward and inverse scenarios. These problems are grounded in Darcy flow, wave propagation, diffusion-reaction, and  Navier-Stokes (NS) equations, which are commonly used in the literature of operator learning and inverse problems~\citep{li2020fourier,lu2022comprehensive,takamoto2022pdebench,iglesias2016bayesian,chada2018parameterizations,zhang2018acoustic}. We summarize the benchmark problems as follows. 
\begin{itemize}
    \item \textbf{D-LINE}: We considered a single-phase 2D Darcy Flow equation. We are interested in predicting from the permeability field to the fluid pressure field (forward) and from pressure field recovering the permeability (inverse). The permeability field is piece-wise constant with a linear interface. 
    \item \textbf{D-CURV}: Similar to D-LINE, but the permeability is piece-wise constant with a curved interface.
    \item \textbf{W-OVAL}: We considered a seismic survey based on an acoustic seismic wave equation~\citep{zhang2018acoustic}. Given an external wave source, the forward task is to predict from the square slowness of the physical media to the wave measurements at the signal receivers. The inverse task is to recover the square slowness from the measurements. The square slowness is piece-wise constant with an oval-shaped interface. 
    \item \textbf{W-Z}: Similar to W-OVAL, except that the square slowness takes a Z-shaped interface.
    \item \textbf{NS}: We considered a viscous, incompressible flow governed by the 2D Navier-Stokes (NS) equation. The forward task is to predict from the initial conditions to the solution at $t=10$, and the inverse task is to recover the initial condition from the solution at $t=10$.
    \item \textbf{DR}: We employed the PDEBench dataset~\citep{takamoto2022pdebench} for a 2D diffusion-reaction system. The forward task is to predict the activator state at time step 5 based on time step 1, while the inverse problem aims to recover the state at step 1 from step 5.
    \item \textbf{CFD}: We used the PDEBench dataset for computational fluid dynamics (CFD) governed by a 2D compressible Navier-Stokes (NS) equation. The forward problem involves predicting the fluid state at time step 2 from the initial state, while the inverse task aims to recover the initial state from time step 2.
    
\end{itemize}
%(1)~\textbf{D-LINE}: We considered a single-phase 2D Darcy Flow equation. We are interested in predicting from the permeability field to the fluid pressure field (forward) and from pressure field recovering the permeability (inverse). The permeability field is piece-wise constant with a linear interface.  (2)~\textbf{D-CURV}: similar to D-LINE, but the permeability is piece-wise constant with a curved interface. (3)~\textbf{W-OVAL}: We considered a seismic survey based on an acoustic seismic wave equation~\citep{zhang2018acoustic}. Given an external wave source, the forward task is to predict from the square slowness of the physical media to the wave measurements at the signal receivers. The inverse task is to recover the square slowness from the measurements. The square slowness is piece-wise constant with an oval-shaped interface.  (4)~\textbf{W-Z}: similar to W-OVAL, except that the square slowness takes a Z-shaped interface. (5)~\textbf{NS}: We considered a 2D Navier-Stokes (NS) equation. The forward task is to predict from the initial conditions to the solution at $t=10$, and the inverse task is to recover the initial condition from the solution at $t=10$. %Following~\citep{li2020fourier,lu2022comprehensive}, we employed relatively small training datasets.
%We employed 800 training and 500 test examples for D-LINE and D-CURV. We used 400 training and 200 test examples for W-OVAL and W-Z. For NS, we used 1000 training and 200 test examples. 
We employed 800 training examples for D-LINE, D-CURV, DR and CFD, 400 for W-OVAL and W-Z, and 1,000 for NS. Each task was evaluated using 200 test examples. 
To access the robustness of our method against data noise and inaccuracy, we conducted additional tests on the first five benchmarks, including D-LINE, D-CURV, W-OVAL, W-Z and NS. We injected 10\% and 20\% white noises into the training dataset. Specifically, for each pair of $\f_n$ and $\u_n$ generated from the numerical solvers, we corrupted them via updating $\f_n \leftarrow \f_n + \eta \bsigma_f \odot {\bepsi_n}$ and $\u_n \leftarrow  \u_n +  \eta \bsigma_u \odot \bxi_n$, where $\eta \in \{0.1, 0.2\}$ represents the noise level, $\bsigma_f$ and $\bsigma_\u$ are the per-element standard deviation of the sampled input and output functions, and $\bepsi_n, \bxi_n \sim \N(\0, \I)$ are Gaussian white noises.
The details of data generation for each benchmark are provided in Appendix Section \ref{sect:data}. 

\begin{table*}
\centering
\caption{\small Relative $L_2$ Error on First Five Benchmarks, where ``D-'' and ``W-''  indicate Darcy flow and wave propagation, respectively, ``LINE'', ``CURV'', ``OVAL'', and ``Z'' represent linear, curved, oval, and z-shaped interfaces, respectively as explained in detail in Appendix Section \ref{sect:exp:darcy} and \ref{sect:exp-wave}; and $0\%, 10\%, 20\%$ indicate the noise level in the training data. N/A indicates absence of reasonable results due to numerical instabilities. }\label{tb:error-final}
\begin{subtable}{\textwidth}
\centering
\caption{\small Forward Prediction}
\begin{tabular}{lccccc}
\toprule
Method &IFNO &IDON &FNO &GNOT \\
\midrule
0\% noise & & & & \\
\toprule
D-LINE &\textbf{3.60e-2} $\pm$ 5.1e-4 &4.71e-1 $\pm$ 3.7e-2 &7.20e-2 $\pm$ 3.5e-3 &3.90e-2 $\pm$ 1.6e-3 \\
D-CURV &3.25e-2 $\pm$ 9.3e-4 &7.68e+0 $\pm$ 3.0e-1 &3.88e-2 $\pm$ 3.1e-4 &\textbf{2.98e-2} $\pm$ 2.0e-4 \\
W-OVAL &\textbf{4.41e-2} $\pm$ 3.2e-3 &3.28e-1 $\pm$ 3.2e-3 &5.52e-2 $\pm$ 1.3e-3 &4.50e-2 $\pm$ 4.0e-3 \\
W-Z &\textbf{3.12e-1} $\pm$ 6.9e-3 &9.88e-1 $\pm$ 9.8e-4 &4.15e-1 $\pm$ 3.3e-3 &N/A \\
NS &1.94e-2 $\pm$ 5.0e-4 &2.70e-1 $\pm$ 1.8e-3 &\textbf{1.54e-2} $\pm$ 8.9e-5 &2.54e-2 $\pm$ 5.0e-4 \\
\midrule
10\% noise & & & & \\
\toprule
D-LINE &\textbf{5.38e-2} $\pm$7.0e-4 &5.12e-1 $\pm$4.2e-2 &8.40e-2 $\pm$6.2e-4&6.90e-2 $\pm$1.2e-3 \\
D-CURV &\textbf{4.96e-2} $\pm$1.5e-3 &7.58e+0 $\pm$5.4e-1 &5.73e-2 $\pm$1.5e-3 &1.50e-1 $\pm$4.1e-2 \\
W-OVAL &6.15e-2 $\pm$7.7e-3 &3.38e-1 $\pm$5.3e-3 &7.88e-2 $\pm$8.9e-4 &\textbf{4.70e-2} $\pm$4.0e-3 \\
W-Z &\textbf{3.13e-1} $\pm$3.1e-2 &9.90e-1 $\pm$2.4e-3 &4.25e-1 $\pm$1.9e-3& N/A \\
NS &\textbf{2.49e-2} $\pm$4.0e-4 &2.75e-1 $\pm$3.9e-3 &2.62e-2 $\pm$2.6e-4& 2.75e-2 $\pm$4.0e-4 \\
\midrule
20\% noise & & & &\\
\toprule
D-LINE &\textbf{7.29e-2} $\pm$2.1e-3 &6.74e-1 $\pm$5.1e-2 &9.96e-2 $\pm$1.5e-3&1.06e-1 $\pm$3.0e-3 \\
D-CURV &\textbf{6.38e-2} $\pm$2.8e-3 &8.85e+0 $\pm$1.4e+0 &7.42e-2 $\pm$5.3e-4&2.07e-1 $\pm$6.8e-2 \\
W-OVAL &7.57e-2 $\pm$4.5e-3 &3.69e-1 $\pm$1.2e-2 &1.20e-1 $\pm$4.3e-3&\textbf{5.42e-2} $\pm$8.0e-3 \\
W-Z &\textbf{3.14e-1} $\pm$7.2e-3 &9.92e-1 $\pm$3.0e-3 &4.42e-1 $\pm$1.8e-3&N/A \\
NS &\textbf{3.48e-2} $\pm$9.7e-4 &2.95e-1 $\pm$1.3e-2 &3.70e-2 $\pm$3.1e-4&2.07e-1 $\pm$6.0e-4 \\
\bottomrule
\end{tabular}
\end{subtable}
\begin{subtable}{\textwidth}
\centering
\caption{\small Inverse  Prediction}
\begin{tabular}{lcccccc}\toprule
Method &IFNO &IDON &NIO &FNO &GNOT \\
\midrule
0\% & & & & &\\
\toprule
D-LINE &\textbf{5.66e-2} $\pm$1.0e-3 &2.98e-1 $\pm$6.9e-2 &2.42e-1 $\pm$5.7e-3 &1.90e-1 $\pm$2.6e-2&2.48e-1 $\pm$1.5e-1 \\
D-CURV &\textbf{5.54e-2} $\pm$9.4e-4 &2.20e-1 $\pm$5.3e-3 &7.83e-2 $\pm$1.4e-3 &6.84e-2 $\pm$5.3e-4&1.19e-1 $\pm$9.0e-3 \\
W-OVAL &\textbf{5.74e-2} $\pm$1.3e-3 &1.42e-1 $\pm$3.4e-3 &8.51e-2 $\pm$6.2e-4 &7.64e-2 $\pm$7.1e-4&7.50e-2 $\pm$1.0e-2 \\
W-Z &\textbf{2.01e-1} $\pm$3.3e-3 &2.61e-1 $\pm$2.2e-3 &2.13e-1 $\pm$9.8e-4 &2.28e-1 $\pm$7.6e-4&4.08e-1 $\pm$6.0e-4 \\
NS &\textbf{5.09e-2} $\pm$6.4e-4 &1.84e-1 $\pm$9.9e-3 &7.07e-2 $\pm$2.6e-4 &5.71e-2 $\pm$8.9e-5&1.10e-1 $\pm$2.7e-3 \\
\midrule
10\% & & & & &\\
\toprule
D-LINE &\textbf{8.99e-2} $\pm$1.4e-3 &3.50e-1 $\pm$4.5e-2 &1.72e-1 $\pm$1.1e-3 &1.63e-1 $\pm$3.8e-3&3.40e-1 $\pm$1.2e-1 \\
D-CURV &\textbf{7.94e-2} $\pm$9.7e-4 &2.25e-1 $\pm$9.3e-3 &2.14e-1 $\pm$3.8e-3 &8.80e-2 $\pm$4.9e-4&2.09e-1 $\pm$9.0e-3 \\
W-OVAL &\textbf{7.19e-2} $\pm$1.6e-3 &1.47e-1 $\pm$9.3e-4 &1.08e-1 $\pm$6.2e-4 &8.49e-2 $\pm$5.8e-4&7.90e-2 $\pm$1.1e-2 \\
W-Z &\textbf{1.91e-1} $\pm$3.1e-3 &2.90e-1 $\pm$5.5e-4 &2.22e-1 $\pm$4.4e-4 &2.30e-1 $\pm$4.4e-4&4.08e-1 $\pm$1.0e-3 \\
NS &\textbf{6.66e-2} $\pm$2.6e-4 &1.84e-1 $\pm$1.8e-2 &1.04e-1 $\pm$6.2e-4 &8.96e-2 $\pm$5.8e-4&1.09e-1 $\pm$8.0e-3 \\
\midrule
20\% & & & & &\\
\toprule
D-LINE &\textbf{1.13e-1} $\pm$9.0e-4 &4.22e-1 $\pm$7.4e-3 &1.89e-1 $\pm$2.3e-3 &1.73e-1 $\pm$5.5e-3&4.53e-1 $\pm$3.7e-2 \\
D-CURV &\textbf{9.58e-2} $\pm$1.2e-3 &2.51e-1 $\pm$1.2e-2 &2.39e-1 $\pm$3.9e-3 &1.07e-1 $\pm$5.3e-4&2.23e-1 $\pm$3.0e-2 \\
W-OVAL &8.85e-2 $\pm$2.6e-3 &1.51e-1 $\pm$4.0e-4 &1.19e-1 $\pm$6.7e-4 &9.71e-2 $\pm$6.7e-4&\textbf{7.92e-2} $\pm$1.1e-2 \\
W-Z &\textbf{2.00e-1} $\pm$3.1e-3 &2.92e-1 $\pm$9.8e-3 &2.27e-1 $\pm$8.9e-4 &2.31e-1 $\pm$1.5e-3&4.09e-1 $\pm$9.0e-4 \\
NS &\textbf{7.52e-2} $\pm$4.7e-4 &1.84e-1 $\pm$2.6e-2 &1.20e-1 $\pm$1.7e-4 &1.16e-1 $\pm$1.6e-3&1.20e-1 $\pm$9.3e-3 \\
\bottomrule
\end{tabular}
\end{subtable}
\end{table*}

\begin{table*}
\centering
\caption{\small Relative $L_2$ Error on Benchmarks DR and CFD.}\label{tb:error-pdebench}\label{tb:pdebench-res}
\begin{subtable}{\textwidth}
\centering
\caption{\small Forward Prediction}
\begin{tabular}{lccccc}
\toprule
Method &IFNO &IDON &FNO &GNOT \\
\toprule
DR &\textbf{9.45e-2} $\pm$ 4.97e-4 &  9.08e-1 $\pm$ 5.83e-5&9.91e-2 $\pm$ 1.34e-4 &6.05e-1 $\pm$ 6.70e-2 \\
CFD &\textbf{1.43e-1} $\pm$ 2.44e-5 &1.55e-1 $\pm$ 5.03e-4 &1.46e-1 $\pm$ 6.26e-4 &1.44e-1 $\pm$ 7.47e-5 \\
\bottomrule
\end{tabular}
\end{subtable}
\begin{subtable}{\textwidth}
\centering
\caption{\small Inverse  Prediction}
\begin{tabular}{lcccccc}\toprule
Method &IFNO &IDON &NIO &FNO &GNOT \\
\toprule
DR &\textbf{2.58e-1} $\pm$9.65e-4 &9.81e-1 $\pm$8.72e-5  &7.42e-1 $\pm$1.22e-2 &3.26e-1 $\pm$8.05e-4&8.83e-1 $\pm$2.22e-2 \\
CFD &\textbf{8.97e-2} $\pm$9.65e-4 & 1.17e-1 $\pm$1.75e-2 &1.09e-1 $\pm$1.44e-3 &9.90e-2 $\pm$2.24e-3&9.08e-2 $\pm$5.55e-5 \\
\bottomrule
\end{tabular}
\end{subtable}
\end{table*}
%\subsection{Results and Analysis}
\noindent\textbf{Methods.} We compared \ours with the invertible Deep Operator Net (\IDON)~\citep{kaltenbach2022semi}, neural inverse operator (NIO)~\citep{molinaro2023neural}, the standard FNO, and a recent state-of-the-art attention-based neural operator model, GNOT~\citep{hao2023gnot}. We evaluated all the methods in both forward and inverse tasks, except for NIO, which is exclusively designed for inverse prediction. In addition, we adapted FNO to  the classical framework~\citep{stuart2010inverse} for solving inverse problems. Specifically, we trained FNO for forward prediction. Then we optimized the input to FNO to match the forward prediction and observations. To leverage the intrinsic structures within the input space, we also used $\beta$-VAE to extract a low-dimensional representation for the input functions values at the sampling locations.  
We optimize the embedding with ADAM. We denote this method as Input Optimization over Forward Model (\clas). Note that we found directly optimizing the function values consistently results in large prediction errors, indicating failure.   
We implemented \ours with PyTorch and used the original implementation of all the competing methods. 

\noindent\textbf{Experimental Settings.} For each approach, we used 100 random examples as the validation set (non-overlapping with the test set) to identify the optimal hyperparameters for each task. The same validation set was used across all the methods for consistency. We employed grid search to select the best hyperparameters, with the specific ranges provided in the Appendix Section~\ref{sect:exp-detail}. Notably, FNO and GNO performed separate validation processes for forward and inverse predictions, obtaining different sets of hyperparameters optimized exclusively for each task. In contrast, \ours and iDON performed validation only once, where the validation error is the sum of the forward and inverse prediction errors, and then they used the same set of hyperparameters to train a single model for both predictions. Following~\citep{lu2022comprehensive},  for each method, we selected the optimal hyperparameters, and then conducted stochastic training for five times, reporting the average relative $L_2$ test error along with the standard deviation. 
%For every approach, we used 100 examples as the validation set (non-overlapping with the test set) to identify the optimal hyperparameters for each task. The validation set is the same for each method in each task. We employed the grid search to select the best hyperparameters. The set of hyperparmaeters and their ranges are given in Appendix. 
%Note that,  FNO and GNOT perform separate validations for forward and inverse prediction. That is, for each type of prediction task, they will use a different set of hyperparameters particularly optimized for that task. In contrast, \ours and iDNOT perform the validation only once (the validation error is the summation of the forward and inverse prediction error), and use a single model with same set of hyperparameters to perform both the forward and inverse prediction. 

% For a fair comparison, we meticulously tuned each method to achieve the optimal performance to the best of our ability. The range of the model size for each method is listed in Appendix Table~\ref{tb:model-size}. 
% The details are provided in Section \ref{sect:exp-detail} of the Appendix. 

%better than other methods, esp FNO, mutual train, IDON --> ablation study
%\noindent\textbf{Predictive Performance.}
\subsection{Predictive Performance}
The relative $L_2$ error of each method is reported in Table \ref{tb:error-final} and~\ref{tb:pdebench-res}. Due to the space limit, the error of IOFM is given in Appendix Table~\ref{tb:iofm}. As shown, in the vast majority of the cases, \ours achieves the best performance, often outperforming the competing approaches by a large margin. In a few cases, \ours is slightly second to GNOT (\eg D-CURV with 0\% noise, and W-OVAL with 10\% noise for forward prediction). This might be mainly caused by FNO itself, which gives 23\%-120\% larger error than GNOT. However, in the forward prediction tasks for W-Z, GNOT consistently encountered numerical issues, and was unable to deliver reasonable prediction errors. Hence we marked its results as N/A. 
\ours greatly surpasses FNO in all the cases, except that in NS forward prediction with 0\% training noise, \ours is slightly worse than FNO. This highlights that our co-learning approach, utilizing a shared FNO architecture, can substantially enhance the performance in both forward and inverse tasks. 

While \IDON can also perform joint forward and inverse predictions, it employs an invertible architecture only in the branch network. For inverse prediction, \IDON needs to back-solve the latent output of the branch network before predicting the input function. This additional step might introduce learning challenges and can substantially increase computational costs. We report the average running time of each method  in  Table \ref{tb:run-time}. 
\begin{table}
\centering
\small 
\caption{\small Average Running Time of Each Method, including both training and prediction, measured at a Linux workstation with NVIDIA GeForce RTX 3090. m: minutes; h: hours.}
\label{tb:run-time}
\begin{tabular}{cccccc}
\hline\hline
\textit{Method}  &  D-LINE & D-CURV & W-OVAL & W-Z & NS\\
\hline 
iFNO & 17m & 68m & 13m & 22m & 50m\\
IDON & 5.7h & 15.3h & 7.8h & 7.9h & 36.8h\\
NIO & 5m & 10m & 5m & 4m & 13m \\
FNO & 11m & 31m & 20m & 15m & 40m\\
GNOT & 23m & 36m & 33m & 29m & 56m\\
IOFM & 99m & 119m & 40m & 35m & 77m\\
\hline\hline
\end{tabular}
\end{table}

\begin{figure*}
\centering
\includegraphics[width=0.65\linewidth]{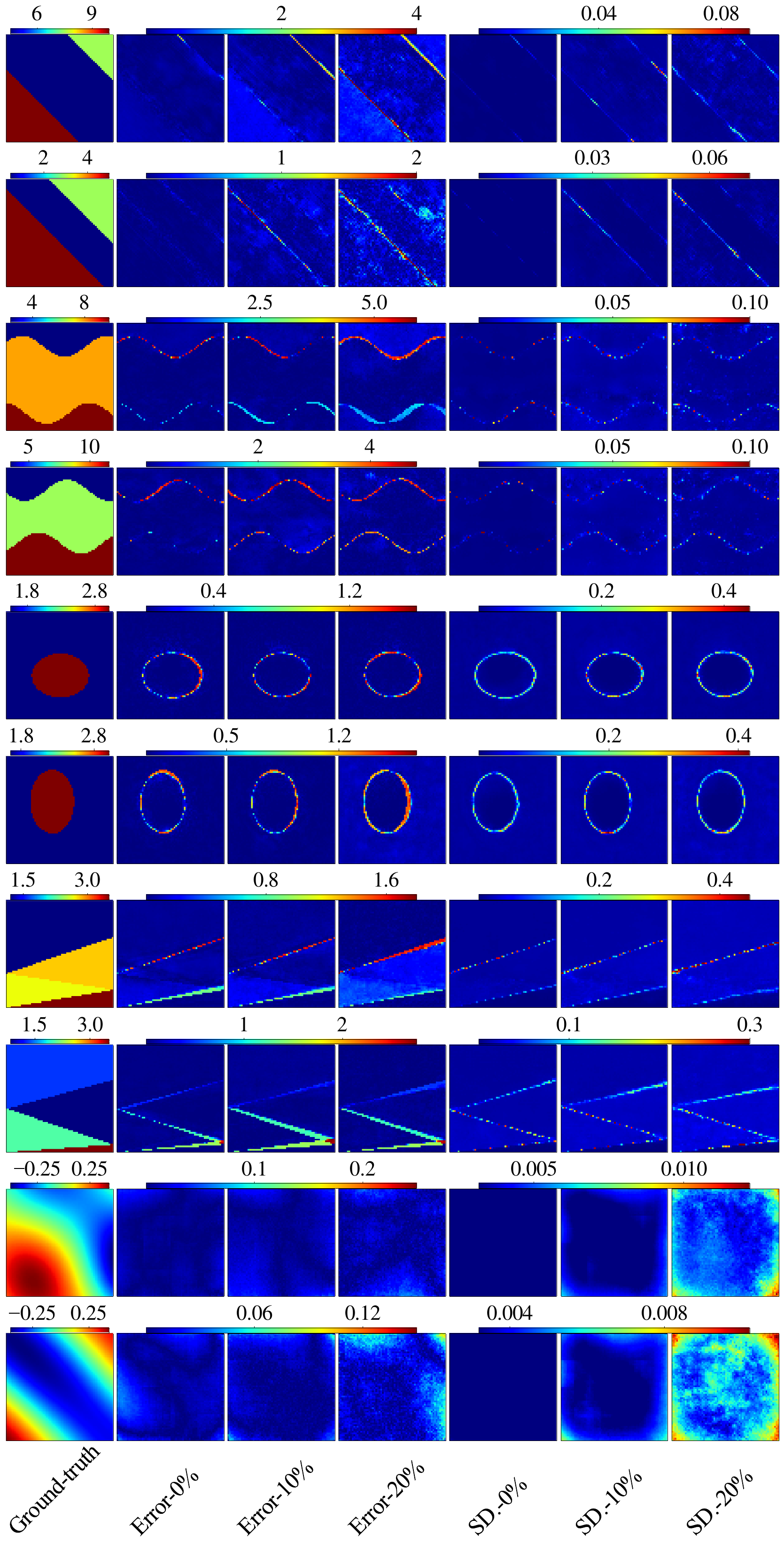}
\caption{\small \ours Pointwise Inverse Prediction Error and Predictive Standard Deviation, denoted by ``Error'' and ``SD'' respectively followed with noise levels in training data. }
\label{fig:point-wise-ours-only}
\end{figure*}

\subsection{Pointwise Error and Prediction Uncertainty}
For a detailed assessment, we performed a fine-grained evaluation by visualizing the pointwise prediction error of \ours in ten randomly selected instances within the inverse scenario. Additionally, we investigated prediction uncertainty by sampling 500 predictions from the VAE component (refer to \eqref{eq:vae}) for each instance. The standard deviation of these predictions at each location is then calculated. The pointwise error is determined based on the predictive mean of the encoder. The outcomes are depicted in Figure \ref{fig:point-wise-ours-only}.

Overall, it is evident that the error grows as the noise level increases, mirroring a similar trend in prediction uncertainty. For problems involving Darcy flow and wave propagation (the first eight instances), the pointwise error is predominantly concentrated at the interfaces between different regions. This pattern is echoed in the prediction uncertainty of \ours, with the standard deviation being significantly larger at the interfaces compared to other locations. These findings are not only intriguing but also intuitively reasonable. Specifically, since the ground-truth permeability and square slowness (that we aim to recover) are identical within the same region but distinct across different regions, predicting values within each region becomes relatively easier, resulting in lower uncertainty (higher confidence). Conversely, at the interfaces, where the ground-truth undergoes abrupt changes, predicting values becomes more challenging, leading to an increase in prediction uncertainty to reflect these challenges (lower confidence).
In the initial condition recovery problem (the last two instances), the pointwise prediction error is primarily concentrated near the boundary of the domain. Correspondingly, the predictive standard deviation of \ours is larger near the boundary. As the noise level increases, the error at the boundary also increases, along with the calibration of uncertainty.

The comparison of pointwise errors with the competing methods is presented in Figure~\ref{fig:point-wise-inverse-comp} and \ref{fig:point-wise-fwd} of the Appendix, encompassing both forward and inverse predictions. The results reveal that competing methods frequently manifest significantly larger errors in various local regions, as illustrated by instances such as \IDON in the second and third instance, NIO in the sixth and seventh instance, and FNO in the seventh and eighth instance in Appendix Fig. \ref{fig:point-wise-inverse-comp}, and \IDON and FNO in nearly all the instances of Appendix Fig. \ref{fig:point-wise-fwd}. We did not include the pointwise error of \clas due to its much inferior performance as compared to all the other methods. 

These results affirm that \ours not only achieves superior global accuracy but also excels in local ground-truth recovery. Furthermore, \ours demonstrates the ability to provide reasonable uncertainty estimates, aligning with the predictive challenges encountered across various local regions.

\noindent\textbf{Ablation Study.} 
%To investigate the impact of each component in \ours, we conducted experiments to assess the model's performance with and without the $\beta$-VAE component. 
%The results are summarized in Table \ref{tb:pretrain}. We can see that while  our model, equipped solely with invertible Fourier blocks, has already substantially surpassed {standard FNO in almost all the cases}, the incorporation of the $\beta$-VAE further enhances the performance in both forward and inverse prediction. In many cases, the improvement is large. For example, it leads to 46\% and 15\% in forward prediction for D-CURV (0\% noise) and NS (20\% noise), and \zhec{fine-tune get worse?}
Finally, we investigated how the performance of \ours varies with the model size. Basically, we found that using three to four invertible Fourier blocks consistently yields optimal performance for \ours.  The detailed results and discussion are provided in  Section \ref{sect:ablation-more} of the Appendix.

%% file: conclusion.tex
%\vspace{-0.1in}
\section{CONCLUSION} 
We have introduced \ours, an invertible Fourier Neural Operator designed to jointly address  forward and inverse problems. By co-learning the bi-directional tasks within a unified architecture, \ours  enhances prediction accuracy for both tasks. Additionally, \ours demonstrates the capability to provide uncertainty calibration for inverse prediction. Through seven benchmark numerical experiments, \ours showcases promising predictive performance. Our architecture  can be easily extended to integrate with other neural operator architectures, such as attention layers. In the future, we plan to explore such extensions, and design a hybrid of exiting or innovative units in the invertible blocks, not necessarily restricted to the Fourier layers. We will continue investigating our method in a broader range of forward prediction and inverse inference tasks.

\section*{Acknowledgements}
This work has been supported by MURI AFOSR grant FA9550-20-1-0358, NSF CAREER Award IIS-2046295, and NSF CSSI-2311685.

%% file: appendix.tex
\section*{Appendix}
\section{Experimental Details}
\subsection{Data Preparation}\label{sect:data}
\subsubsection{Darcy Flow}\label{sect:exp:darcy}
A single-phase 2D Darcy Flow equation was employed,
\begin{align} 
	-\nabla \cdot( a(\x) \nabla u(\x))&=g(\x)\quad \x \in (0,1)^2  \notag \\
    u(\x)&= 0, \quad \x \in \partial(0,1)^2,  \label{eq:darcy}
\end{align}
where $a(\x)$ is the permeability field, $u(\x)$ is fluid pressure, and $g(\x)$ is an external source. We considered a practically useful case where the permeability is piece-wise constant across separate regions $\{\Rcal_i\}^C_{i=1}$ where $\Rcal_1 \cup \ldots \cup \Rcal_C = (0, 1)^2$, and $$a(\x)=\sum_{i=1}^{C}q_i\mathbf{1}_{\Rcal_i}(\x).$$ For the forward scenario, we are interested in predicting the pressure field $u(\x)$ based on the given permeability field $a(\x)$. Conversely, in the inverse scenario, the goal is to recover $a(\cdot)$ from the measurement of the pressure field $u(\cdot)$. We considered two benchmark problems, each featuring the permeability with a different type of geometric structures. For both problems, we fixed $g(x)$ to $1$.

\begin{figure*}
	\centering
		\begin{subfigure}[b]{0.33\textwidth}
			\centering
			\includegraphics[width=\textwidth]{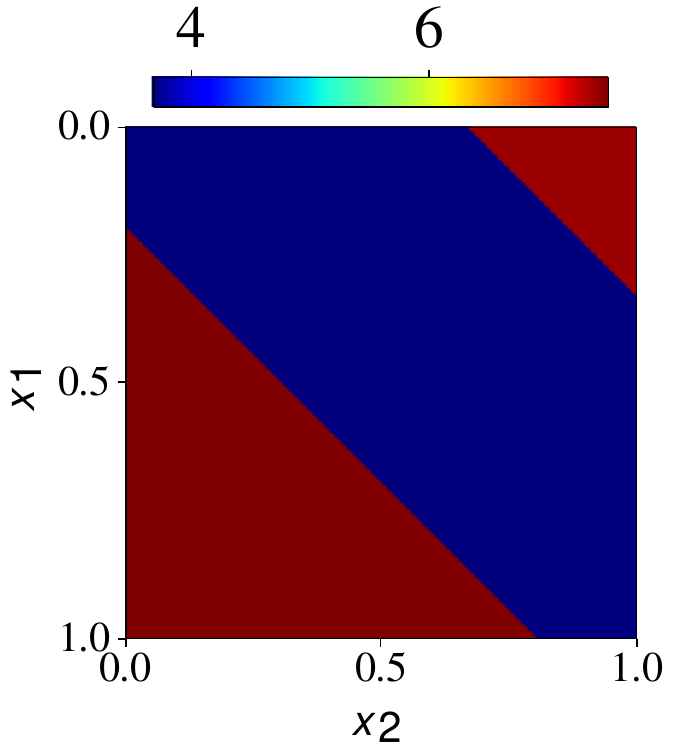}
			\caption{\small \textit{Permeability}}
		\end{subfigure}
		\begin{subfigure}[b]{0.308\textwidth}
			\centering
			\includegraphics[width=\textwidth]{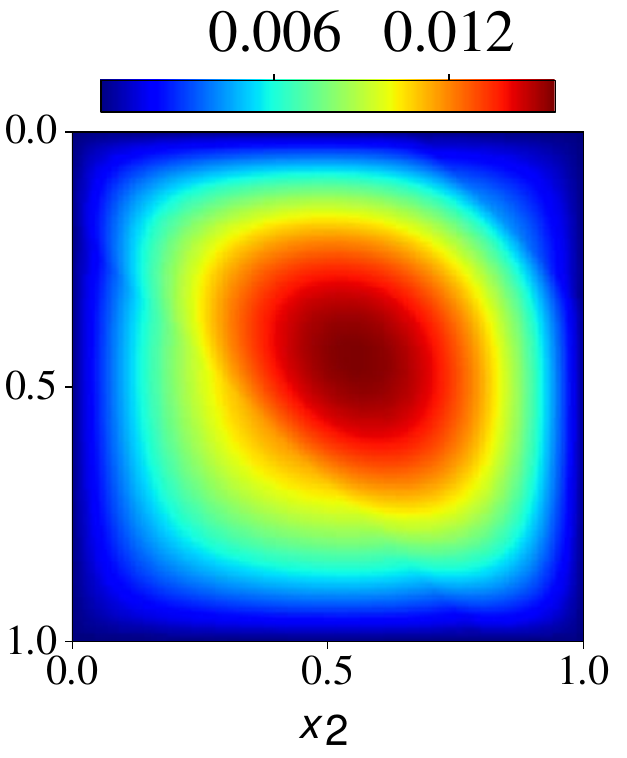}
			\caption{\small \textit{Solution}}
		\end{subfigure}
	%\vspace{-0.1in}
	\caption{\small Darcy Flow with Piece-Wise Permeability and Linear Interfaces.} \label{fig:dline-example}
	%\vspace{-0.13in}
\end{figure*}

\textbf{Linear permeability interface (D-LINE)}. In the first problem, the domain was divided into three regions (namely $C=3$), and the interfaces were represented by parallel lines inclined at a 45-degree angle from the horizontal axis; see Figure \ref{fig:dline-example} for an example. The permeability value in each region was sampled from a uniform distribution $U(0, 10)$. To determine the position of these sections, we sampled the endpoints of interface lines, denoted by $\{(w_1, 0), (0, w_1)\}$ for the first interface, and $\{(w_2, 0),(0, w_2)\}$ for the second interface, respectively, where $w_1 \sim \Ucal(0, \frac{1}{3})$ and $w_2 \sim \Ucal(\frac{1}{3}, \frac{2}{3})$. 

\begin{figure*}
	\centering
		\begin{subfigure}[t]{0.33\textwidth}
			\centering
			\includegraphics[width=\textwidth]{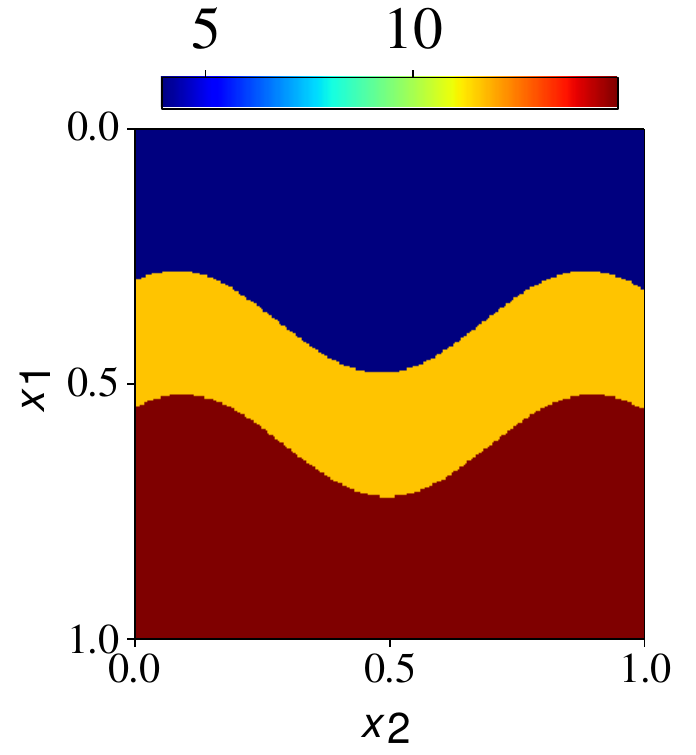}
			\caption{\small \textit{Permeability}}
		\end{subfigure}
		\begin{subfigure}[t]{0.32\textwidth}
			\centering
			\includegraphics[width=\textwidth]{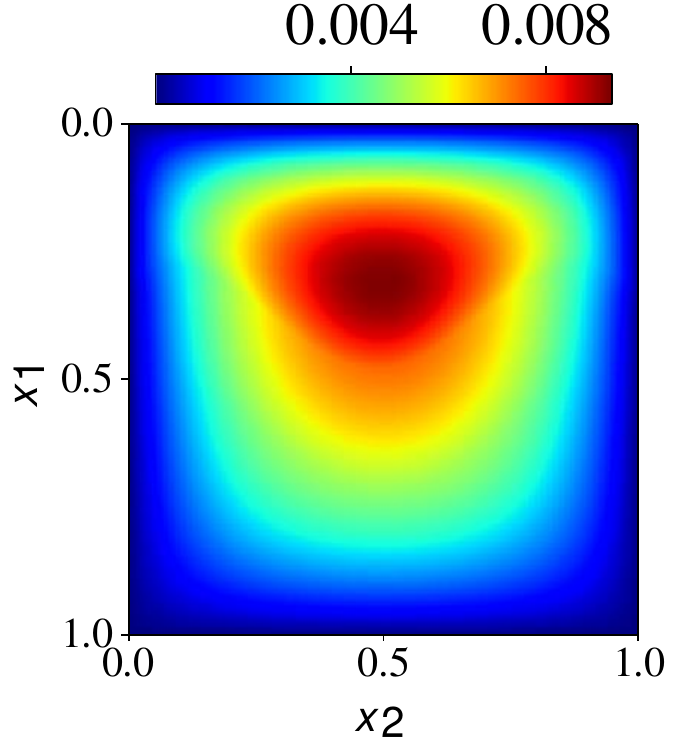}
			\caption{\small \textit{Solution}}
		\end{subfigure}

	%\vspace{-0.1in}
	\caption{\small Darcy Flow with Piece-Wise Permeability and Curved Interfaces.}\label{fig:dcurv-example}
	%\vspace{-0.13in}
\end{figure*}

\textbf{Curved permeability interface (D-CURV)}. In the second problem, the domain was also partitioned into three regions, but the interfaces were  curves; see Figure \ref{fig:dcurv-example} for an illustration. We sampled the first interface as $x_2=p_1+0.1\sin(2.5\pi(x_1+r_1))$ and the second $x_2=p_2+0.1\sin(2.5\pi(x_1+r_2))$, where $p_1\sim \mathcal{U}(0.15,0.4), p_2\sim \mathcal{U}(0.6,0.85)$, and $r_1,r_2\sim \mathcal{U}(0,1)$. The permeability value in each region was sampled from $U(0, 15)$.

To prepare the dataset, we followed~\citep{li2020fourier} to apply a second-order finite difference solver %on a $253 \times 253$ grid, and downsampled the 
and collected pairs of
permeability and pressure field on a $64 \times 64$ grid. 
%For the case of linear permeability interfaces,  we used 300 training examples, while for the curved interfaces, we used 800 training examples. In both problems, we evaluated our method with 500 test examples.  
To assess the robustness of our method against data noise and inaccuracy, we conducted tests by
injecting 10\% and 20\% white noises into the training datasets, as described in the main paper. 
%Specifically, for each pair of $\f_n$ and $\u_n$ generated by the simulator, we corrupted them via updating $\f_n \leftarrow \f_n + \eta \bsigma_f \odot {\bepsi_n}$ and $\u_n \leftarrow  \u_n +  \eta \bsigma_u \odot \bxi_n$, where $\eta \in \{0.1, 0.2\}$ represents the noise level, $\bsigma_f$ and $\bsigma_\u$ are the per-element standard deviation of the sampled input and output functions, and $\bepsi_n, \bxi_n \sim \N(\0, \I)$ are Gaussian white noises. 
We provide the  signal-to-noise ratios in Table \ref{tb:snr}.

\subsubsection{Wave Propagation}\label{sect:exp-wave}
\begin{figure*}[t]
	\centering
	\setlength\tabcolsep{0pt}
	\includegraphics[width=\textwidth]{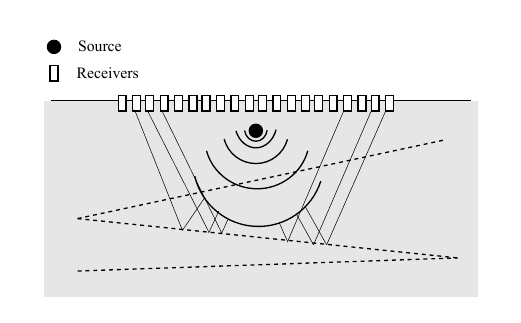}
	\caption{\small An Illustration of The Seismic Surveys, where the shaded region is the physical media of interest and dashed lines are the interface for different regions of square slowness.} \label{fig:wave propagation}
\end{figure*}

We employed an acoustic seismic wave equation to simulate seismic surveys,
\begin{align} 
&m(\x) \frac{\d^2 u(\x,t)}{\d t^2} - \nabla^2 u(\x,t) + \eta \frac{\d u(x,t)}{\d t}=q(t), \;\x \in \Omega,\notag \\
&u(\x, 0) =  \frac{\d u(x,0)}{\d t} = 0,  \x \in \Omega, \;\;u(\x, t) = 0, \x \in \partial \Omega, \notag 
\end{align}
%with zero initial conditions $u(x,0) = 0 $ and $\frac{d u(x,0)}{dt} = 0$, and Dirichlet conditions $u(x,t)= 0,$ $x\in \partial\Omega$. Here 
where $m(\x)$ is square slowness, defined as the inverse of squared wave speed in the given physical media, $q(t)$ expresses the external wave source, and $\eta$ is the damping mask. In the simulation, the physical media was placed in the domain $\Omega = (0,1.28\unit{km})^2$, where  $\forall \x=(x_1, x_2) \in \Omega$, $x_1$ and $x_2$ represent the depth and width, respectively. For each survey, an external source was positioned at particular location to initiate seismic waves. A row of receivers were placed at a particular depth to record the wave measurements across time; see Figure \ref{fig:wave propagation} for an illustration.
We conducted simulation over a duration of one second. In the forward tasks, we intended to predict the wave measurements at the receivers based on the square slowness $m(\x)$, while for the inverse tasks, the goal was to recover $m(\x)$ from the measurements at the receivers. We designed two benchmark problems, each corresponding to a different class of structures for $m(\x)$. In both problems, we set $q(t) = (1-2\pi^2 f_0^2 (t - \frac{1}{f_0})^2 )e^{- \pi^2 f_0^2 (t - \frac{1}{f_0})}$, where $f_0=0.01$. This implies that  the  peak frequency of the source wave is 10 Hz. 

\begin{figure*}
	\centering
		\begin{subfigure}[t]{0.33\textwidth}
			\centering
			\includegraphics[width=\textwidth]{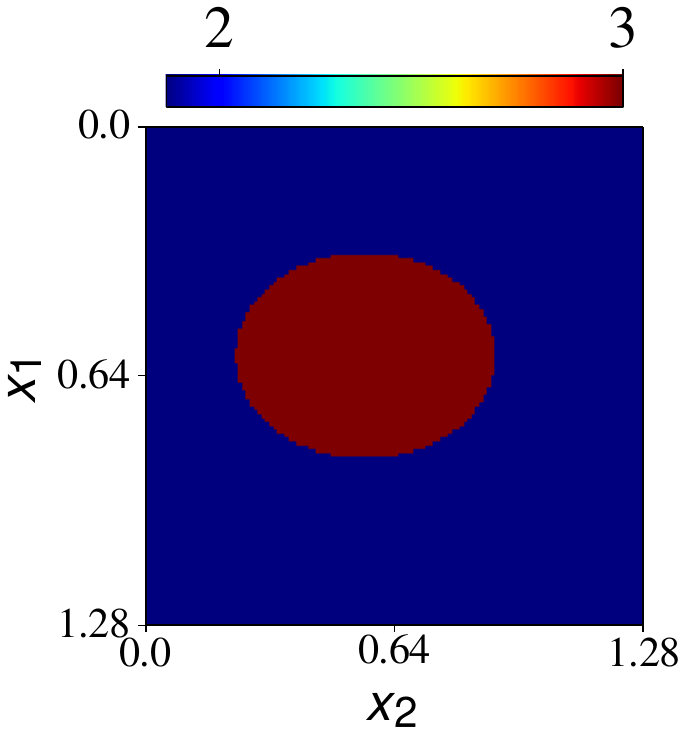}
			\caption{\small \textit{Square Slowness}}
		\end{subfigure}
		\begin{subfigure}[t]{0.32\textwidth}
			\centering
			\includegraphics[width=\textwidth]{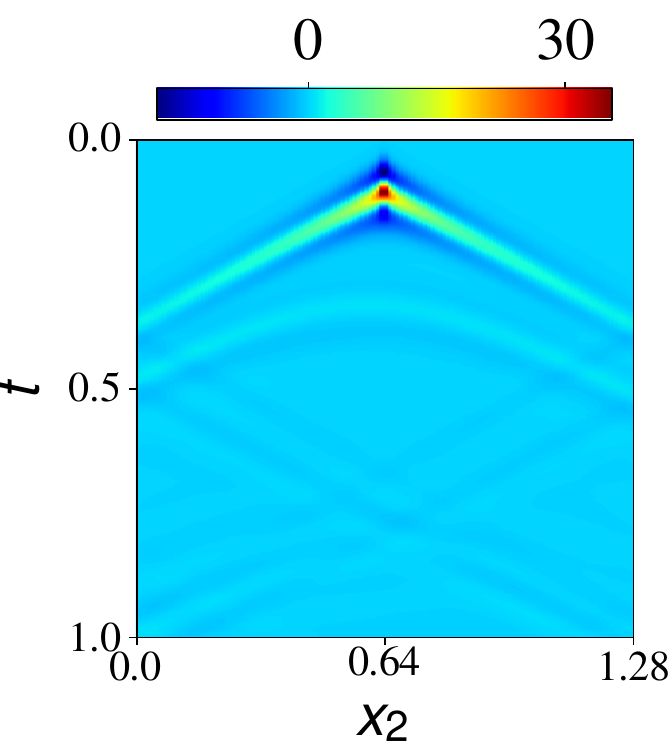}
			\caption{\small \textit{Measurements at Receivers}}
		\end{subfigure}
	%\vspace{-0.1in}
	\caption{\small Wave Propagation via Piece-Wise Square Slowness and Oval-Shaped Interface.}\label{fig:w-oval-example}
	%\vspace{-0.13in}
\end{figure*}

\textbf{Oval-shaped square slowness (W-OVAL)}.
In the first problem, the source was placed at a depth of 50m and horizontally in the middle. The receivers were positioned at a depth of 20m. The domain was partitioned into two regions, with $m(\x)$ being the same within each region. See Figure \ref{fig:w-oval-example} for an example and the surveyed data by the receivers. The interface is an oval, defined by its center $(x_{1c}, x_{2c})$, and two radii $w$ and $h$.  We sampled $x_{1c},x_{2c} \sim \mathcal{U}(\frac{1}{4}\times 1.28\unit{km},\frac{3}{4} \times 1.28\unit{km})$,  and $w, h\sim \mathcal{U}(\frac{1}{10}\times 1.28\unit{km},\frac{1}{5}\times 1.28\unit{km})$. Inside the ellipse, we set the value of $m$ to 3, while the outside the ellipse, we sampled the value from $\Ucal(1, 2)$. We used the Devito library\footnote{\url{https://github.com/devitocodes/devito}} to simulate the receivers' measurements. The square slowness was generated on a $128 \times 128$ grid, and the measurements were computed at $614$ time steps. The data was then downsampled to a $64 \times 64$ grid and $62$ time steps.

\begin{figure*}[t]
	\centering
		\begin{subfigure}[t]{0.33\textwidth}
			\centering
			\includegraphics[width=\textwidth]{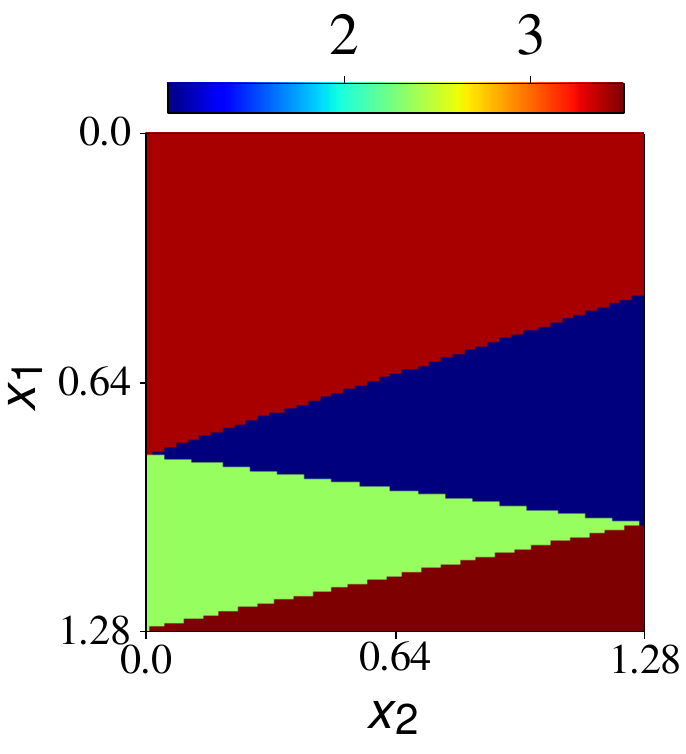}
			\caption{\small \textit{Square Slowness}}
		\end{subfigure}
		\begin{subfigure}[t]{0.33\textwidth}
			\centering
			\includegraphics[width=\textwidth]{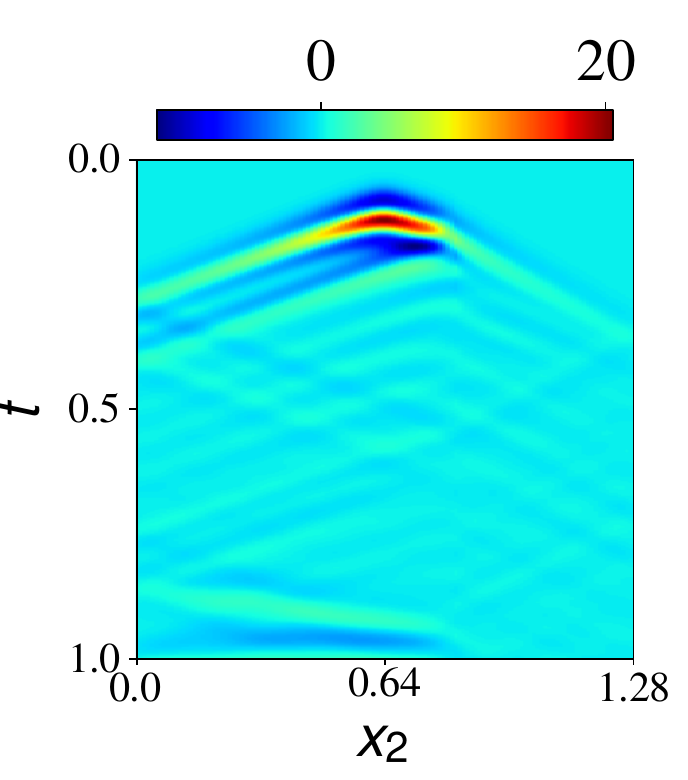}
			\caption{\small \textit{Measurements at Receivers}}
		\end{subfigure}

	%\vspace{-0.1in}
	\caption{\small Wave Propagation via Piece-Wise Square Slowness and Z-Shaped Interface.}\label{fig:w-z-example}
	%\vspace{-0.13in}
\end{figure*}

\textbf{Z-shaped square slowness (W-Z)}.
In the second problem, the source was positioned at a depth of 80m (still horizontally in the middle). The domain was divided into four regions, with the interfaces between these regions forming a z-shape. The  values of $m(\x)$ inside each region are identical; see Figure \ref{fig:w-z-example} for an example. We represented the end points of these interfaces by $(1.28\unit{km},0)$, $(z_1,1.28\unit{km})$, $(z_2,0)$, and $(z_3,1.28\unit{km})$, where $z_1 \sim U(\frac{3}{4}\times 1.28\unit{km},1.28\unit{km})$, $z_2 \sim U(\frac{1}{2}\times 1.28\unit{km},\frac{3}{4}\times1.28\unit{km})$, and $z_3 \sim \mathcal{U}(\frac{1}{4}\times 1.28\unit{km},\frac{1}{2}\times1.28\unit{km})$. 
We set $m=3.5$ in the bottom region, and sampled the value of $m(\x)$ for each of the other three regions from $\Ucal(0.5, 3.5)$. The receivers' measurements were computed at 573 time steps with spatial resolution $128$, and then downsampled at 58 steps with spatial resolution 64.

\subsubsection{Navier-Stoke Equation (NS)} \label{sect:exp:wave}
\begin{figure*}[t]
	\centering
		\begin{subfigure}[t]{0.33\textwidth}
			\centering
			\includegraphics[width=\textwidth]{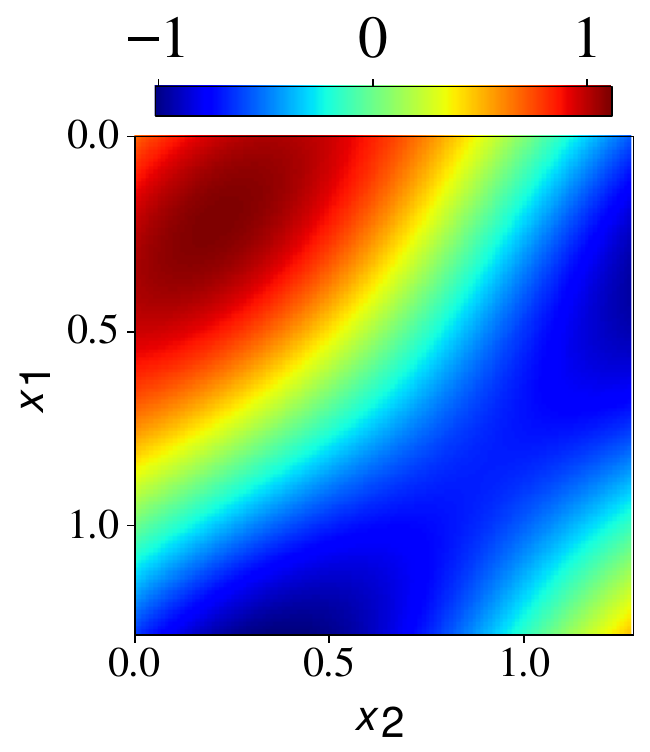}
			\caption{\small \textit{Initial Condition}}
		\end{subfigure}
		\begin{subfigure}[t]{0.33\textwidth}
			\centering
			\includegraphics[width=\textwidth]{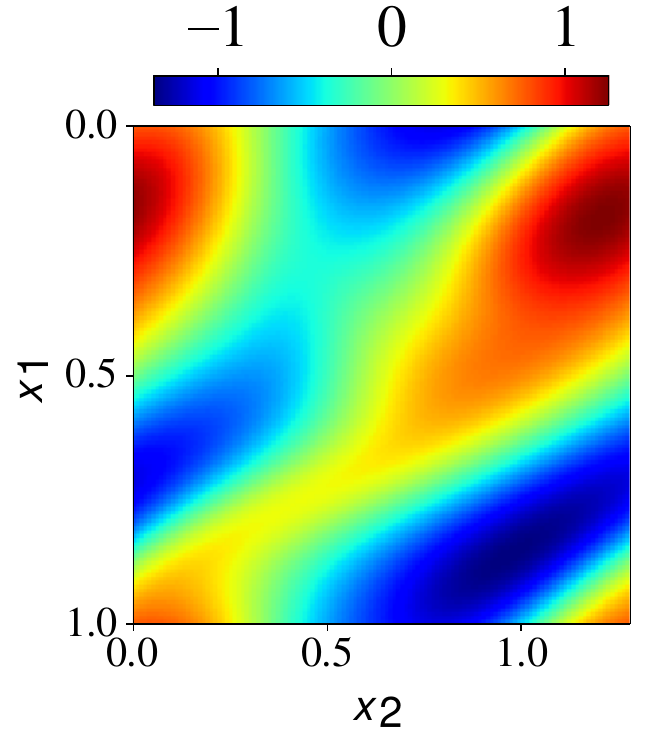}
			\caption{\small \textit{Solution at } $t=10$}
		\end{subfigure}

	%\vspace{-0.1in}
	\caption{\small Solution of NS Equation.}\label{fig:ns-example}
	%\vspace{-0.13in}
\end{figure*}

We considered the 2D Navier-Stokes (NS) equation as used in~\citep{li2020fourier}.  
The solution represents the vorticity of a viscous, incompressible fluid within the spatial domain $\x = (x_1, x_2) \in [0, 1]^2$. The viscosity was set to $10^{-3}$. In the forward scenario, we aim to predict the vorticity at time $t=10$ from the initial condition. Correspondingly, in the inverse scenario, the goal is to reconstruct the initial condition $\omega_0$ from the observed vorticity at $t=10$. We generated the initial condition by $\omega_0(x_1,x_2)=\sum_{i=1}^{2} \sum_{j=1}^{2} q_{ij}\sin(\alpha_i\pi (x_1+c_j)) \cdot q_{ij}\cos(\alpha_i\pi (x_2+c_j))$, where $\alpha_i\sim \mathcal{U}(0.5,1), c_j\sim \mathcal{U}(0,1)$, and $q_{ij}\sim \mathcal{U}(-1,1)$. An example is given by Figure \ref{fig:ns-example}. 
For each experiment, we used $1000$ training examples and $200$ test examples generated at a $64 \times 64$ grid. Again, we performed additional tests by injecting 10\% and 20\% noises into both the training input and output function samples.  

\subsubsection{Diffusion Reaction (DR)}
We employed a PDEBench dataset for a 2D diffusion reaction system with two non-linearly coupled variables,
\begin{align}
    \partial_t u &= D_u \partial_{xx} u + D_u \partial_{yy} u + R_u, \\
    \partial_t v &= D_v \partial_{xx} v + D_v \partial_{yy} v + R_v, 
\end{align}
where $x,y\in(-1,1)$, $u=u(t,x,y)$ is the activator, $v=v(t,x,y)$ is the inhibitor, and the diffusion coefficients $D_u = 1\times10^{-3}$ and $D_v = 5\times 10^{-3}$. The activator and inhibitor are coupled via the Fitzhugh-Nagumo equation,
\begin{align}
    R_u(u,v) &= u - u^3 - k - v, \\
    R_v(u,v) &= u - v.
\end{align}
The initial condition is generated from a standard normal distribution. We extracted the original dataset on a $64 \times 64$ grid.

\subsubsection{Computational Fluid Dynamics (CFD)}
We used a CFD dataset from PDEBench. The governing equation is a 2D Compressible Navier-Stokes equation,
\begin{align}
    \partial_t \rho + \nabla \cdot (\rho \mathbf{v}) &= 0, \\
    \rho (\partial_t \mathbf{v} + \mathbf{v} \cdot \nabla \mathbf{v}) &= -\nabla p + \eta \Delta \mathbf{v} + (\zeta + \eta/3) \nabla (\nabla \cdot \mathbf{v}), \\
    \partial_t \left[ \epsilon + \frac{\rho v^2}{2} \right] 
    &+ \nabla \cdot \left[ \left( \epsilon + p + \frac{\rho v^2}{2} \right) \mathbf{v} - \mathbf{v} \cdot \boldsymbol{\sigma}' \right] = 0,
\end{align}
where $\rho, \mathbf{v}, p, \epsilon$ are the mass density, velocity, gas pressure, and internal energy, respectively. We used the simulation data with Mach number $M=\frac{|v|}{c_s}=0.1$, where $c_s$ is  the sound velocity, and with shear and bulk viscosity $\eta = \zeta = 0.1$. %The goal was to predict the solution at timestamp 2 from the solution at timestamp 0, and the inverse task was to predict the solution at timestamp 0 from the solution at timestamp 2. We used 800 training examples and 200 test examples.

\subsection{Methods}
\label{sect:exp-detail}
All the models were trained with AdamW or the Adam optimizer with exponential decay strategy or the reducing learning rate on plateau strategy. 
%All the models were trained with Adam optimizer. 
The learning rate was chosen from $\{10^{-5}, 10^{-4}, 5 \times 10^{-4}, 10^{-3}, 10^{-2}\}$.
We varied the mini-batch size from \{10, 20, 50\}. %For each method, we performed cross-validation on the training set to select the model architecture and other hyperparameters.  
\begin{itemize}
	\item 
    \textbf{\ours}. %We used four invertible Fourier blocks for all the experiments.
    %We set the lifting dimension to 64 (channel numbers). 
    We selected the number of invertible Fourier blocks from \{1, 2, 3, 4\}, the channel lifting dimension from \{32, 64, 128\}, the number of Fourier modes (for frequency truncation) from \{8, 12, 16, 32\}, the number of training epochs for invertible Fourier blocks  from \{100, 200, 500\}, the number of training epochs for $\beta$-VAE from
 \{200, 500, 1000\}, and the number of joint training  epochs was set from \{50, 100, 500\}. The architecture of $\beta$-VAE is the same accross all the benchmarks. We employed a Gaussian encoder that includes five convolutional layers with 32, 64, 128, 256 and 512 channels respectively.  
   % We maintained a consistent architecture for the VAE modules across all benchmarks.  We applied five convolutional layers to increasingly lift the number of channels to 32, 64, 128, 256, and 512. 
   The decoder first applies five transposed convolutional layers to sequntially reduce the number of channels to 512, 256, 128, 64, and 32. Then a transposed convolutional layer (with 32 output channels) and another convolutional layer (with one output channel) are applied to produce the prediction of $\f$. We set $\beta =0.01$ for all the benchmarks except for Darcy flow, we set $\beta=10^{-6}$.
        \item 
    \textbf{FNO}. We used the original FNO implementation (\url{https://github.com/neuraloperator/neuraloperator}).
    To ensure convergence, we set the number of training epochs  to 1000. The lifting dimension was chosen from $\{32,64,128\}$. The number of Fourier modes was tuned from $\{8,12,16,32\}$. The number of Fourier layers was selected from $\{1, 2,3,4\}$.
    \item 
    \textbf{\IDON}. We used the Jax implementation of \IDON from the authors (\url{https://github.com/pkmtum/Semi-supervised_Invertible_Neural_Operators/tree/main}). The number of training epochs was set to 1000. We varied the number of layers for the branch net and trunk set from \{1, 3, 4, 5\}.
    Note that to ensure invertibility, the width of the branch net must be set to the dimension of the (discretized) input function. For instance, if the input function is sampled on a $64 \times 64$ grid, the width of the branch net will be $4096$. The inner-width of the trunk net was selected from \{5, 10, 20\} and the width of the branch net. 
     For a fair comparison, only the forward and backward supervised loss terms were retained for training. 
    
        \item 
    \textbf{NIO.} We used the PyTorch implementation from the authors (\url{https://github.com/mroberto166/nio}). We employed 1000 training epochs. NIO used convolution layers for the branch net of the deepONet module. We tuned the number of convolution layers from \{6, 8, 10\}, and kernel size from \{3, 5\}, and padding from \{1, 3\}. The stride is fixed to 2. For the trunk net, we tuned the number of layers from $\{2,3,4\}$, and the layer width from  $\{32,64,128,256\}$. For the FNO component, we varied the number of Fourier layers from $\{1,2,3, 4\}$,  lifting dimension from $\{32,64,128\}$, and the number of Fourier modes from $\{8,12,16,32\}$. 
    \item 
    {\textbf{\clas.}} We used the original FNO implementation from the authors (\url{https://github.com/neuraloperator/neuraloperator}) to train the forward model. The hyper-parameter selection and training follow the same way as we used FNO for forward and backward prediction.   We used the same $\beta$-VAE as in \ours. To conduct inverse prediction, 
    we ran 1000 ADAM epochs optimize the input embeddings. The step-size for the optimization was selected from $\{10^{-3}, 10^{-2}, 10^{-1}, 0.2, 0.5\}$
%    \item 
%    \textbf{CNO.} We used the original CNO(2d) implementation from the authors (\url{https://github.com/camlab-ethz/ConvolutionalNeuralOperator}) to train the forward and backward model. During hyper-parameter tuning, we vary the number of res block from $\{1, 2, 3\}$, number of res neck from $\{4, 6, 8\}$ and channel multiplier from $\{16, 32, 64\}$. All other hyper-parameters are set to default as appeared in their repository. We ran ADAM for 1000 epochs with learning rate of 1E-4. 
    \item 
    \textbf{GNOT.} We used the original implementation from authors (\url{https://github.com/HaoZhongkai/GNOT}) to train forward and inverse prediction models. For hyper-parameter selection, we varied the dimension of the embeddings from $\{64, 128, 256\}$ and number of attention layers from $\{2, 3, 4, 5\}$. We used the default data normalization(unit) as in their repository. 
    The model was trained with the default optimizer(ADAMW), with weight decay of 5E-6, gradient clip of 0.999, with 50 warm up epochs and 500 training epochs in total. 
 \end{itemize}

\begin{table}
\caption{\small {The Signal-To-Noise Ratios (SNR) in DB for Datasets With Noises.}}
\small
\centering
\begin{tabular}{ccccccccc}
\hline \multirow{2}{*}{ \textit{Benchmark} } & \multicolumn{2}{c}{Forward Prediction}  & \multicolumn{2}{c}{Inverse Prediction} \\
\cline { 2 - 3 } \cline { 4 - 5 } & 10\% noise& 20\% noise   & 10\% noise & 20\% noise \\
\hline 
D-LINE & 22.2 &16.1   & 28.4   & 22.4\\
D-CURV &20.1  &  14.1    &28.5  & 22.5\\
W-OVAL & 20.0 &13.9 & 32.4   & 26.4\\
W-Z &20.0  & 13.9    & 29.7 & 23.6\\
NS & 20.7 &14.7  &  21.4  & 15.4\\
\hline
\end{tabular}
\label{tb:snr}
 \end{table}
% \begin{table}
% \small
% \centering
% \begin{tabular}{cccccc}
% \hline 
%  & \ours & IDON & NIO & FNO & \clas\\
% \hline
% Min. & 2M  & 16M & 2M  &4M &2M \\
% Max. & 18M  & 117M & 25M & 36M &18M \\
% \hline
% \end{tabular}
% \caption{\small  {The model size range for each method.} Note that the smallest IDON employs one feed-forward layer in the branch-net, which takes 16M parameters (the input dimension to the branch-net is at least $64\times 64=4906$ in our experiments). For FNO, we consider the total size of two separate models, one for forward prediction and the other for inverse prediction. }
% \label{tb:model-size}
% \end{table}

\begin{figure*}
\centering
\includegraphics[width=0.75\linewidth]{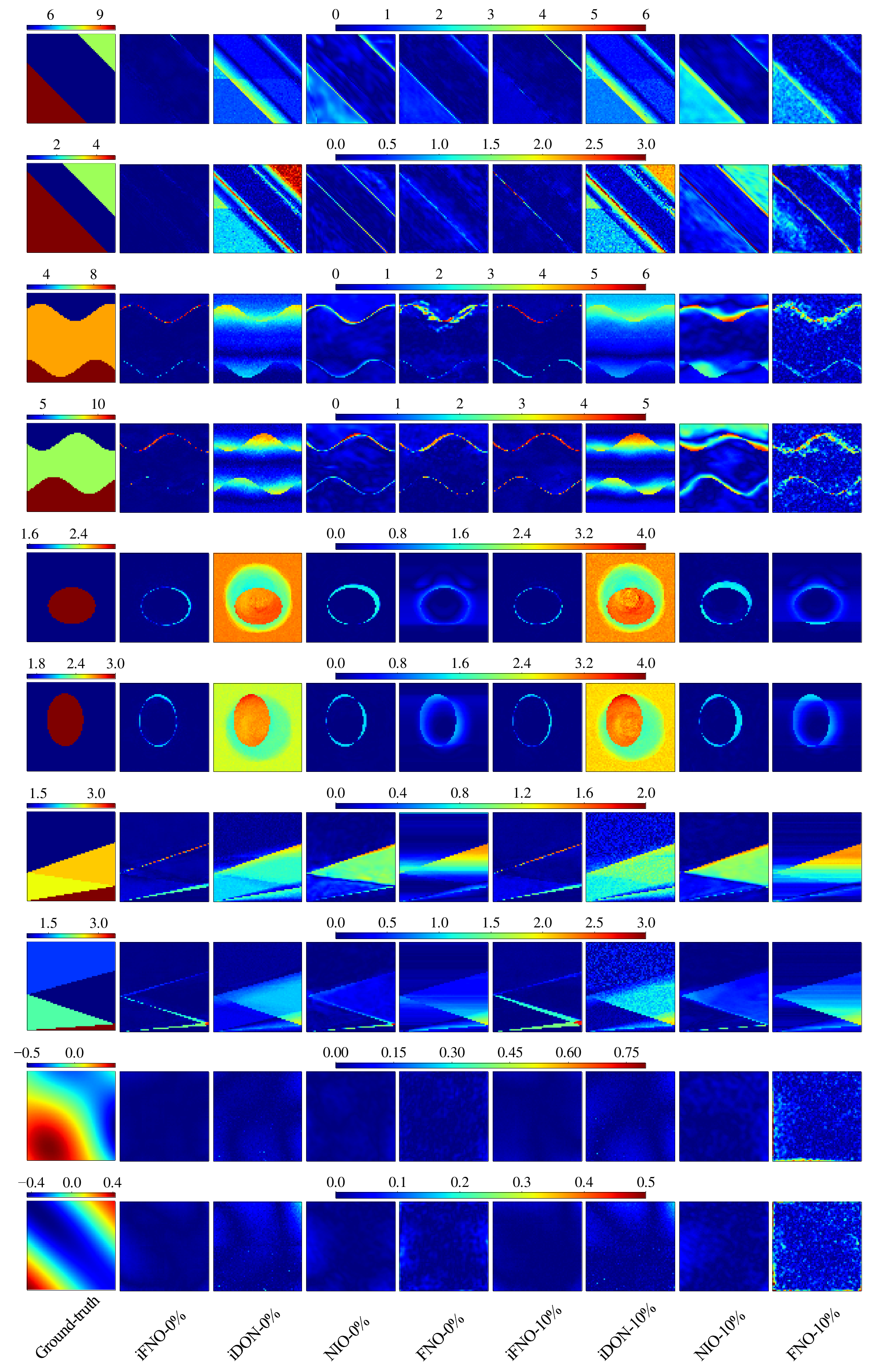}
\caption{\small Pointwise Error of Inverse Prediction, where ``-0\%'' and ``-10\%'' indicate the noise level in the training data.}
\label{fig:point-wise-inverse-comp}
\end{figure*}

\begin{figure*}
\centering
\includegraphics[width=0.67\linewidth]{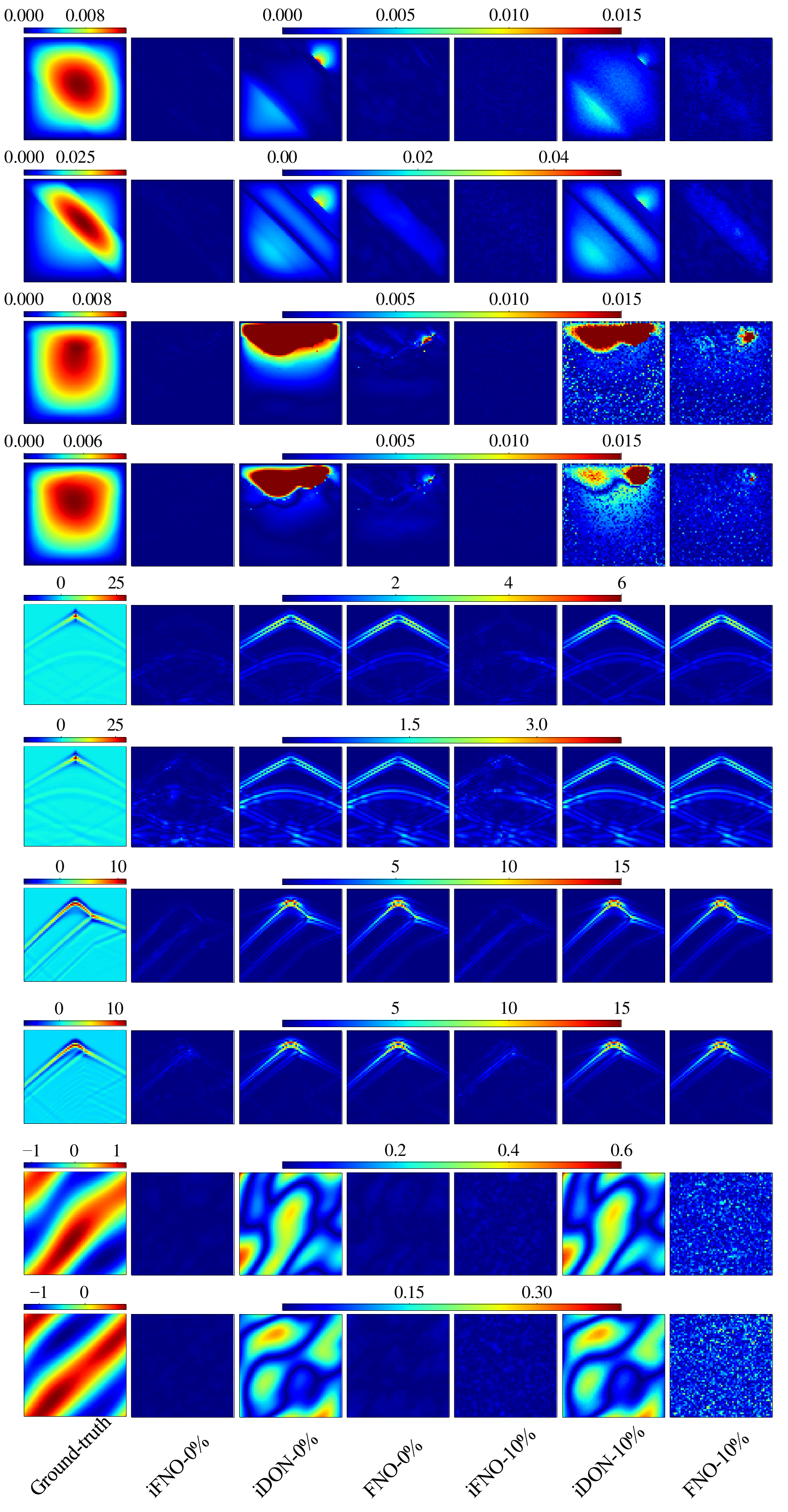}
\caption{\small Pointwise Error of Forward Prediction, where ``-0\%'' and ``-10\%'' indicate the noise level in the training data.}
\label{fig:point-wise-fwd}
\end{figure*}

% \begin{table*}[htbp!]
% \small
% \centering
% \caption{\small Running Time of Each Method, including both training and prediction, measured at a Linux workstation with NVIDIA GeForce RTX 3090.}
% \label{tb:run-time}
% \begin{tabular}{cccccccc}
% \hline\hline
% \textit{Benchmark}  &  \ours & IDON & NIO & FNO & {\clas} & GNOT\\
% \hline 
% % small-10 & 4 & $22.88 \%$ & & 3 & $29.97 \%$ \\
% D-LINE & 17 mins  &5.7 hours  &5 mins  &11 mins & {99 mins} & 23 mins\\
% D-CURV &68 mins   &15.3 hours  &10 mins  &31 mins & {119 mins} & 36 mins\\
% W-OVAL &13 mins  &7.8 hours  & 5 mins &20 mins & {40 mins} & 33 mins\\
% W-Z &22 mins   &7.9 hours  & 4 mins & 15 mins& {35 mins} & 29 mins\\
% NS &50 mins  &36.8 hours  &  13 mins &40 mins & {77 mins} & 56 mins\\
% \hline
% \end{tabular}
% \end{table*}

\section{Ablation Study}\label{sect:ablation-more}
\begin{figure*}
	\centering
		\begin{subfigure}[t]{0.41\textwidth}
			\centering
			\includegraphics[width=\textwidth]{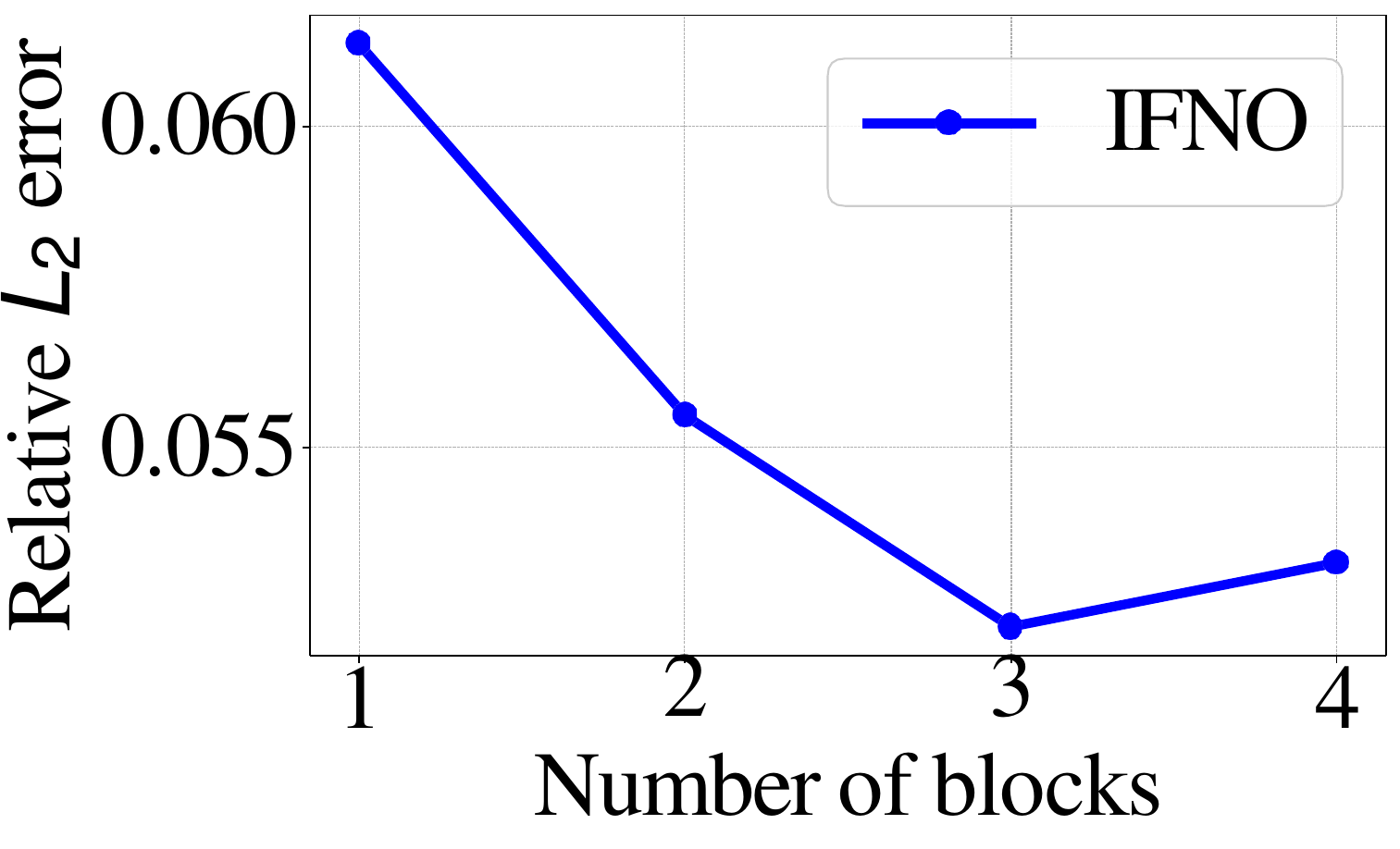}
			\caption{\small \textit{Forward}}
		\end{subfigure}
		\begin{subfigure}[t]{0.42\textwidth}
			\centering
			\includegraphics[width=\textwidth]{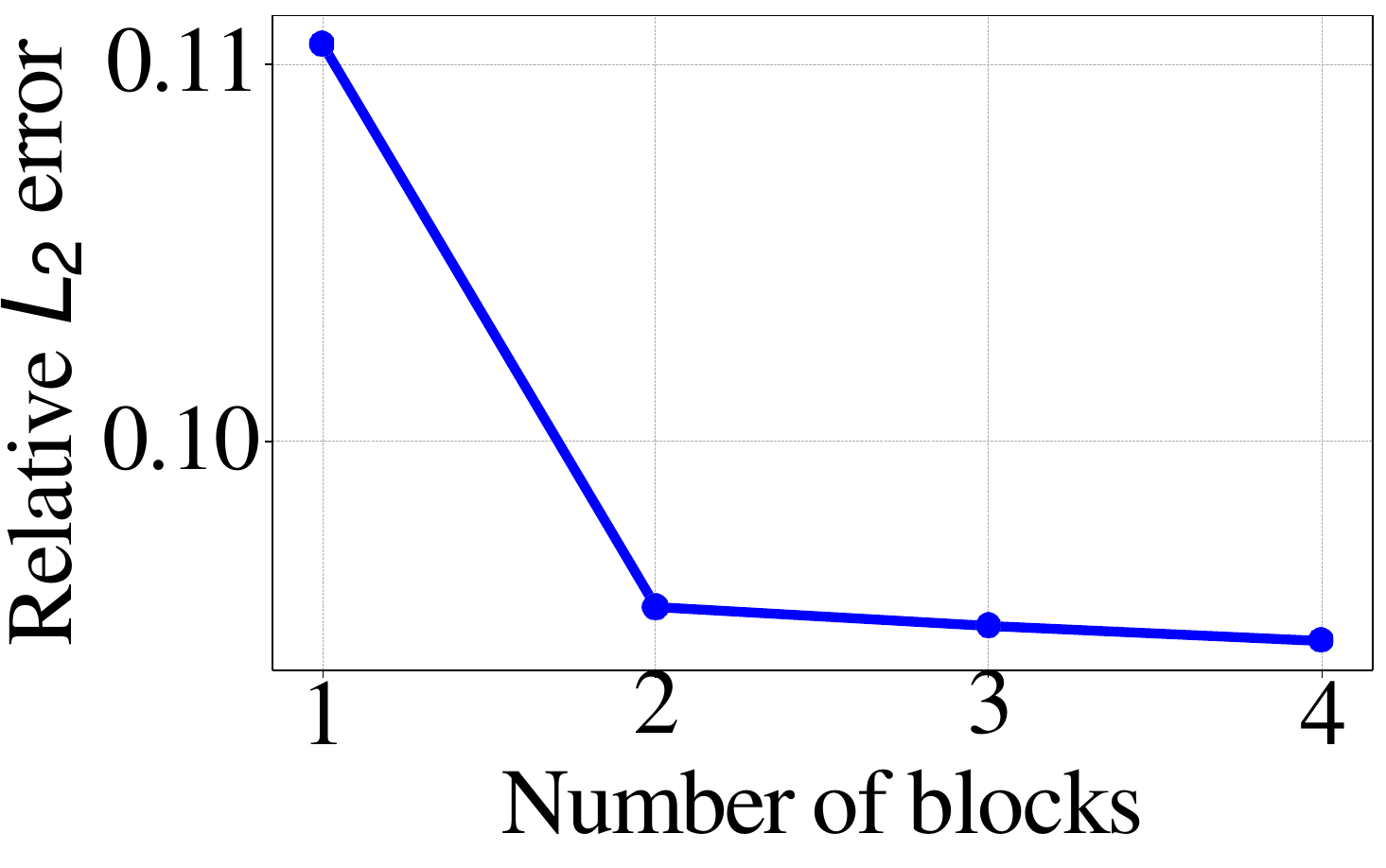}
			\caption{\small \textit{Inverse}}
		\end{subfigure}

	%\vspace{-0.1in}
	\caption{\small Relative $L_2$ Error of \ours on D-LINE with 10\% Noise in Training Data.}\label{fig:ablation-dline}
	%\vspace{-0.13in}
\end{figure*}

\begin{figure*}
	\centering
		\begin{subfigure}[t]{0.41\textwidth}
			\centering
			\includegraphics[width=\textwidth]{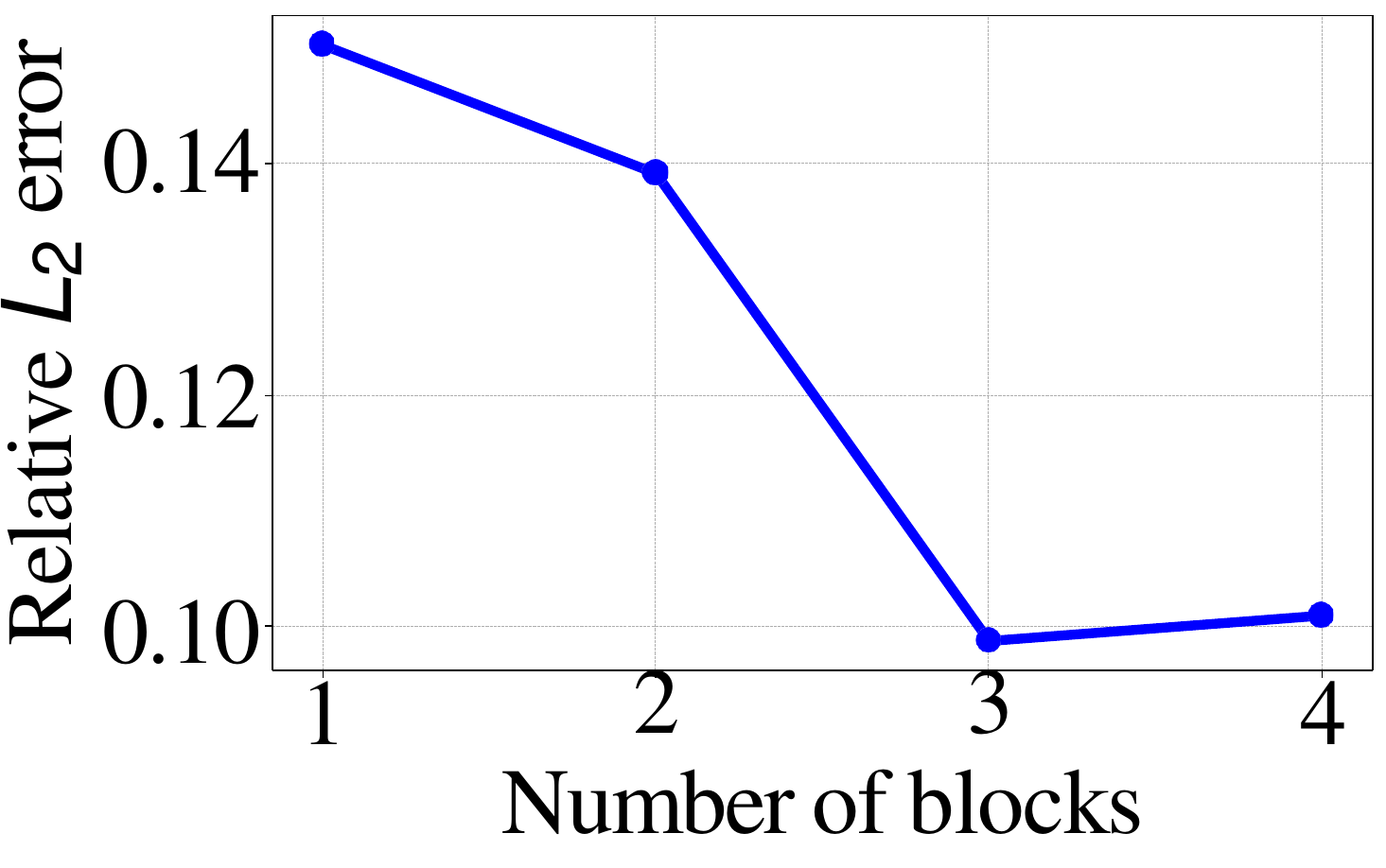}
			\caption{\small \textit{Forward}}
		\end{subfigure}
		\begin{subfigure}[t]{0.42\textwidth}
			\centering
			\includegraphics[width=\textwidth]{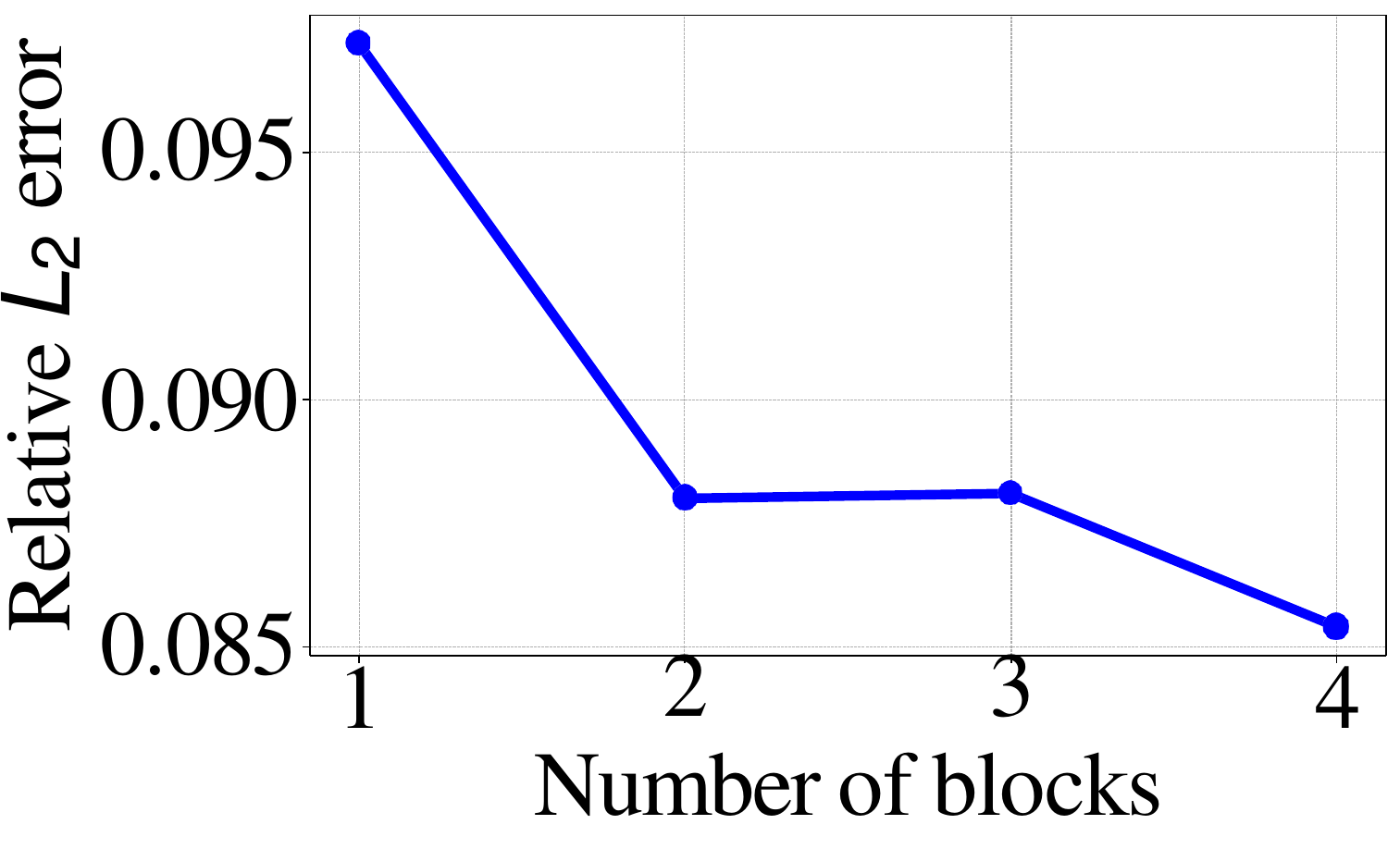}
			\caption{\small \textit{Inverse}}
		\end{subfigure}

	%\vspace{-0.1in}
	\caption{\small Relative $L_2$ Error of \ours on W-OVAL with 10\% Noise in Training Data.}\label{fig:ablation-woval}
	%\vspace{-0.13in}
\end{figure*}

We further investigate how the performance of \ours varies along with model size. To this end, we ran \ours on D-LINE and W-OVAL,  with 10\% noise level in the training data.   For D-LINE, we fixed the number of Fourier modes to 32 and the channel lifting dimension to 64. For W-OVAL, we fixed the number of Fourier modes to 8 and the channel lifting dimension to 64. We varied the number of invertible Fourier blocks from \{1, 2, 3, 4\}. The test relative $L_2$ error for  the forward and inverse prediction is shown in Fig.~\ref{fig:ablation-dline} and Fig.~\ref{fig:ablation-woval}. As we can see, %\ours consistently outperforms all the competing methods, even only using one invertible Fourier block. 
with more blocks, the prediction accuracy of \ours can be further improved. The best performance is achieved at three and four blocks for D-LINE and W-OVAL, respectively, in terms of the average error for the forward and inverse prediction.

\begin{table*}
\centering
\caption{\small Relative $L_2$ Error on Each Inverse Prediction Task with IOFM.}\label{tb:iofm}
%\begin{subtable}{.45\linewidth}
\centering
%\caption{\small Inverse  Prediction}
\begin{tabular}{lcc}\toprule
Method &IFNO & IOFM\\
\midrule
0\% & & \\
\toprule
D-LINE &\textbf{5.66e-2} $\pm$1.0e-3 & 5.25e-1 $\pm$ 3.13e-2\\
D-CURV &\textbf{5.54e-2} $\pm$9.4e-4 & 1.78e-1 $\pm$ 4.61e-3\\
W-OVAL &\textbf{5.74e-2} $\pm$1.3e-3 & 2.91e-1 $\pm$ 2.33e-2\\
W-Z &\textbf{2.01e-1} $\pm$3.3e-3 & 7.42e-1 $\pm$ 5.96e-2\\
NS &\textbf{5.09e-2} $\pm$6.4e-4 & 5.78e-2 $\pm$ 1.79e-3\\
\midrule
10\% & & \\
\toprule
D-LINE &\textbf{8.99e-2} $\pm$1.4e-3 & 4.61e-1 $\pm$ 2.05e-2\\
D-CURV &\textbf{7.94e-2} $\pm$9.7e-4 & 2.17e-1 $\pm$ 3.40e-3\\
W-OVAL &\textbf{7.19e-2} $\pm$1.6e-3 & 2.29e-1 $\pm$ 1.12e-2\\
W-Z &\textbf{1.91e-1} $\pm$3.1e-3 & 7.02e-1 $\pm$ 4.38e-2\\
NS &\textbf{6.66e-2} $\pm$2.6e-4 & 9.02e-2 $\pm$ 1.79e-3\\
\midrule
20\% & & \\
\toprule
D-LINE &\textbf{1.13e-1} $\pm$9.0e-4 & 3.61e-1 $\pm$ 2.14e-2\\
D-CURV &\textbf{9.58e-2} $\pm$1.2e-3 & 2.35e-1 $\pm$ 8.14e-3\\
W-OVAL &\textbf{8.85e-2} $\pm$2.6e-3 & 2.01e-1 $\pm$ 7.02e-3\\
W-Z &\textbf{2.00e-1} $\pm$3.1e-3 & 8.53e-1 $\pm$ 7.96e-2\\
NS &\textbf{7.52e-2} $\pm$4.7e-4 & 2.36e-1 $\pm$ 6.22e-3\\
\bottomrule
\end{tabular}
\end{table*}

%% file: IFNO.bbl
\begin{thebibliography}{}

\bibitem[Batlle et~al., 2023]{batlle2023kernel}
Batlle, P., Darcy, M., Hosseini, B., and Owhadi, H. (2023).
\newblock Kernel methods are competitive for operator learning.
\newblock {\em arXiv preprint arXiv:2304.13202}.

\bibitem[Cao, 2021]{cao2021choose}
Cao, S. (2021).
\newblock Choose a transformer: {F}ourier or galerkin.
\newblock {\em Advances in neural information processing systems}, 34:24924--24940.

\bibitem[Chada et~al., 2018]{chada2018parameterizations}
Chada, N.~K., Iglesias, M.~A., Roininen, L., and Stuart, A.~M. (2018).
\newblock Parameterizations for ensemble kalman inversion.
\newblock {\em Inverse Problems}, 34(5):055009.

\bibitem[Dinh et~al., 2016]{dinh2016density}
Dinh, L., Sohl-Dickstein, J., and Bengio, S. (2016).
\newblock Density estimation using real nvp.
\newblock {\em arXiv preprint arXiv:1605.08803}.

\bibitem[Engl and Groetsch, 2014]{engl2014inverse}
Engl, H.~W. and Groetsch, C.~W. (2014).
\newblock {\em Inverse and ill-posed problems}, volume~4.
\newblock Elsevier.

\bibitem[Gupta et~al., 2021]{gupta2021multiwavelet}
Gupta, G., Xiao, X., and Bogdan, P. (2021).
\newblock Multiwavelet-based operator learning for differential equations.
\newblock {\em Advances in neural information processing systems}, 34:24048--24062.

\bibitem[Hao et~al., 2023]{hao2023gnot}
Hao, Z., Wang, Z., Su, H., Ying, C., Dong, Y., Liu, S., Cheng, Z., Song, J., and Zhu, J. (2023).
\newblock Gnot: A general neural operator transformer for operator learning.
\newblock In {\em International Conference on Machine Learning}, pages 12556--12569. PMLR.

\bibitem[Higgins et~al., 2016]{higgins2016beta}
Higgins, I., Matthey, L., Pal, A., Burgess, C., Glorot, X., Botvinick, M., Mohamed, S., and Lerchner, A. (2016).
\newblock beta-vae: Learning basic visual concepts with a constrained variational framework.
\newblock In {\em International conference on learning representations}.

\bibitem[Iglesias et~al., 2016]{iglesias2016bayesian}
Iglesias, M.~A., Lu, Y., and Stuart, A.~M. (2016).
\newblock A {B}ayesian level set method for geometric inverse problems.
\newblock {\em Interfaces and free boundaries}, 18(2):181--217.

\bibitem[Kaltenbach et~al., 2022]{kaltenbach2022semi}
Kaltenbach, S., Perdikaris, P., and Koutsourelakis, P.-S. (2022).
\newblock Semi-supervised invertible deeponets for bayesian inverse problems.
\newblock {\em arXiv preprint arXiv:2209.02772}.

\bibitem[Kaltenbach et~al., 2023]{kaltenbach2023semi}
Kaltenbach, S., Perdikaris, P., and Koutsourelakis, P.-S. (2023).
\newblock Semi-supervised invertible neural operators for bayesian inverse problems.
\newblock {\em Computational Mechanics}, pages 1--20.

\bibitem[Kashefi and Mukerji, 2024]{kashefi2024novel}
Kashefi, A. and Mukerji, T. (2024).
\newblock A novel fourier neural operator framework for classification of multi-sized images: Application to 3d digital porous media.
\newblock {\em arXiv preprint arXiv:2402.11568}.

\bibitem[Kingma and Welling, 2013]{kingma2013auto}
Kingma, D.~P. and Welling, M. (2013).
\newblock Auto-encoding variational {B}ayes.
\newblock {\em arXiv preprint arXiv:1312.6114}.

\bibitem[Kovachki et~al., 2023]{kovachki2023neural}
Kovachki, N., Li, Z., Liu, B., Azizzadenesheli, K., Bhattacharya, K., Stuart, A., and Anandkumar, A. (2023).
\newblock Neural operator: Learning maps between function spaces with applications to pdes.
\newblock {\em Journal of Machine Learning Research}, 24(89):1--97.

\bibitem[Li et~al., 2024]{li2024multi}
Li, S., Yu, X., Xing, W., Kirby, R., Narayan, A., and Zhe, S. (2024).
\newblock Multi-resolution active learning of fourier neural operators.
\newblock In {\em International Conference on Artificial Intelligence and Statistics}, pages 2440--2448. PMLR.

\bibitem[Li et~al., 2020]{li2020fourier}
Li, Z., Kovachki, N.~B., Azizzadenesheli, K., Bhattacharya, K., Stuart, A., Anandkumar, A., et~al. (2020).
\newblock Fourier neural operator for parametric partial differential equations.
\newblock In {\em International Conference on Learning Representations}.

\bibitem[Li et~al., 2022]{li2022transformer}
Li, Z., Meidani, K., and Farimani, A.~B. (2022).
\newblock Transformer for partial differential equations' operator learning.
\newblock {\em arXiv preprint arXiv:2205.13671}.

\bibitem[Long et~al., 2022]{long2022kernel}
Long, D., Mrvaljevic, N., Zhe, S., and Hosseini, B. (2022).
\newblock A kernel approach for pde discovery and operator learning.
\newblock {\em arXiv preprint arXiv:2210.08140}.

\bibitem[Lu et~al., 2021]{lu2021learning}
Lu, L., Jin, P., Pang, G., Zhang, Z., and Karniadakis, G.~E. (2021).
\newblock Learning nonlinear operators via deeponet based on the universal approximation theorem of operators.
\newblock {\em Nature machine intelligence}, 3(3):218--229.

\bibitem[Lu et~al., 2022]{lu2022comprehensive}
Lu, L., Meng, X., Cai, S., Mao, Z., Goswami, S., Zhang, Z., and Karniadakis, G.~E. (2022).
\newblock A comprehensive and fair comparison of two neural operators (with practical extensions) based on fair data.
\newblock {\em Computer Methods in Applied Mechanics and Engineering}, 393:114778.

\bibitem[Molinaro et~al., 2023]{molinaro2023neural}
Molinaro, R., Yang, Y., Engquist, B., and Mishra, S. (2023).
\newblock Neural inverse operators for solving pde inverse problems.
\newblock {\em arXiv preprint arXiv:2301.11167}.

\bibitem[Nashed and Scherzer, 2002]{nashed2002inverse}
Nashed, M.~Z. and Scherzer, O. (2002).
\newblock {\em Inverse Problems, Image Analysis, and Medical Imaging: AMS Special Session on Interaction of Inverse Problems and Image Analysis, January 10-13, 2001, New Orleans, Louisiana}, volume 313.
\newblock American Mathematical Soc.

\bibitem[Pathak et~al., 2022]{pathak2022fourcastnet}
Pathak, J., Subramanian, S., Harrington, P., Raja, S., Chattopadhyay, A., Mardani, M., Kurth, T., Hall, D., Li, Z., Azizzadenesheli, K., et~al. (2022).
\newblock Fourcastnet: A global data-driven high-resolution weather model using adaptive fourier neural operators.
\newblock {\em arXiv preprint arXiv:2202.11214}.

\bibitem[Pfaff et~al., 2020]{pfaff2020learning}
Pfaff, T., Fortunato, M., Sanchez-Gonzalez, A., and Battaglia, P.~W. (2020).
\newblock Learning mesh-based simulation with graph networks.
\newblock {\em arXiv preprint arXiv:2010.03409}.

\bibitem[Raonic et~al., 2023]{raonic2024convolutional}
Raonic, B., Molinaro, R., De~Ryck, T., Rohner, T., Bartolucci, F., Alaifari, R., Mishra, S., and de~B{\'e}zenac, E. (2023).
\newblock Convolutional neural operators for robust and accurate learning of pdes.
\newblock {\em Advances in Neural Information Processing Systems}, 36.

\bibitem[Ronneberger et~al., 2015]{ronneberger2015u}
Ronneberger, O., Fischer, P., and Brox, T. (2015).
\newblock U-net: Convolutional networks for biomedical image segmentation.
\newblock In {\em Medical image computing and computer-assisted intervention--MICCAI 2015: 18th international conference, Munich, Germany, October 5-9, 2015, proceedings, part III 18}, pages 234--241. Springer.

\bibitem[Stuart, 2010]{stuart2010inverse}
Stuart, A.~M. (2010).
\newblock Inverse problems: a {B}ayesian perspective.
\newblock {\em Acta numerica}, 19:451--559.

\bibitem[Takamoto et~al., 2022]{takamoto2022pdebench}
Takamoto, M., Praditia, T., Leiteritz, R., MacKinlay, D., Alesiani, F., Pfl{\"u}ger, D., and Niepert, M. (2022).
\newblock Pdebench: An extensive benchmark for scientific machine learning.
\newblock {\em Advances in Neural Information Processing Systems}, 35:1596--1611.

\bibitem[Tanaka and Dulikravich, 1998]{tanaka1998inverse}
Tanaka, M. and Dulikravich, G.~S. (1998).
\newblock {\em Inverse problems in engineering mechanics}.
\newblock Elsevier.

\bibitem[Tikhonov, 1963]{tikhonov1963solution}
Tikhonov, A.~N. (1963).
\newblock On the solution of ill-posed problems and the method of regularization.
\newblock In {\em Doklady akademii nauk}, volume 151, pages 501--504. Russian Academy of Sciences.

\bibitem[Yilmaz, 2001]{yilmaz2001seismic}
Yilmaz, {\"O}. (2001).
\newblock {\em Seismic data analysis: Processing, inversion, and interpretation of seismic data}.
\newblock Society of exploration geophysicists.

\bibitem[Zhang et~al., 2018]{zhang2018acoustic}
Zhang, W., Joardar, A.~K., et~al. (2018).
\newblock Acoustic based crosshole full waveform slowness inversion in the time domain.
\newblock {\em Journal of Applied Mathematics and Physics}, 6(05):1086.

\bibitem[Zhao et~al., 2022]{zhao2022learning}
Zhao, Q., Lindell, D.~B., and Wetzstein, G. (2022).
\newblock Learning to solve pde-constrained inverse problems with graph networks.
\newblock {\em arXiv preprint arXiv:2206.00711}.

\end{thebibliography}
